\documentclass[10pt,journal,compsoc]{IEEEtran}
\usepackage{amsmath,amsfonts}
\usepackage{array}
\usepackage[caption=false,font=normalsize,labelfont=sf,textfont=sf]{subfig}
\usepackage{textcomp}
\usepackage{stfloats}
\usepackage{url}
\usepackage{verbatim}
\usepackage{xcolor} 

\usepackage{graphicx}
\usepackage{cite}
\usepackage[square,numbers]{natbib}
\usepackage[colorlinks,citecolor=red,urlcolor=blue,bookmarks=false,hypertexnames=true]{hyperref}

\usepackage{balance}
\usepackage{xcolor}

\definecolor{ruby}{rgb}{0.88, 0.07, 0.37}
\definecolor{tealblue}{rgb}{0.18, 0.40, 0.46}
\usepackage{graphicx}%
\usepackage{multirow}%
\usepackage{amsmath,amssymb,amsfonts, bm}%
\usepackage{mathrsfs}%
\usepackage{xcolor}%
\usepackage{textcomp}%
\usepackage{manyfoot}%
\usepackage{booktabs}%
\usepackage{listings}%
\usepackage{bbm}
\usepackage{balance}
\usepackage{wrapfig,lipsum}
\usepackage{subfig}
\usepackage{mathtools}
\DeclarePairedDelimiterX{\infdivx}[2]{(}{)}{%
  #1\;\delimsize\|\;#2%
}

\usepackage{mathabx}
\usepackage{algorithm}
\usepackage{algpseudocode}
\algrenewcommand\algorithmicrequire{\textbf{Input:}}
\algrenewcommand\algorithmicensure{\textbf{Output:}}

\newtheorem{definition}{Definition}

\newtheorem{remark}{Remark}

\begin{document}

\title{Open-CRB: Towards Open World \\ Active Learning for 3D Object Detection}

\author{Zhuoxiao Chen, Yadan Luo, Zixin Wang, Zijian Wang, Zi Huang~\IEEEmembership{Senior Member,~IEEE} 
\thanks{The authors are with the School of Electrical Engineering and Computer Science, St Lucia, QLD 4072, Australia \\(e-mail: zhuoxiao.chen@uq.edu.au; y.luo@uq.edu.au; zixin.wang@uq.edu.au; zijian.wang@uq.edu.au; 
helen.huang@uq.edu.au).} 
}

\markboth{Journal of \LaTeX\ Class Files,~Vol.~14, No.~8, August~2021}%
{Shell \MakeLowercase{\textit{et al.}}: A Sample Article Using IEEEtran.cls for IEEE Journals}

\IEEEpubid{0000--0000/00\$00.00~\copyright~2021 IEEE}

\maketitle

\begin{abstract} LiDAR-based 3D object detection has recently seen significant advancements through active learning (AL), attaining satisfactory performance by training on a small fraction of strategically selected point clouds. However, in real-world deployments where streaming point clouds may include unknown or novel objects, the ability of current AL methods to capture such objects remains unexplored. This paper investigates a more practical and challenging research task: Open World Active Learning for 3D Object Detection (OWAL-3D), aimed at acquiring informative point clouds with new concepts. To tackle this challenge, we propose a simple yet effective strategy called Open Label Conciseness (OLC), which mines novel 3D objects with minimal annotation costs. Our empirical results show that OLC successfully adapts the 3D detection model to the open world scenario with just a single round of selection. Any generic AL policy can then be integrated with the proposed OLC to efficiently address the OWAL-3D problem. Based on this, we introduce the Open-CRB framework, which seamlessly integrates OLC with our preliminary AL method, CRB, designed specifically for 3D object detection. We develop a comprehensive codebase for easy reproducing and future research, supporting 15 baseline methods (\textit{i.e.}, active learning, out-of-distribution detection and open world detection), 2 types of modern 3D detectors (\textit{i.e.}, one-stage SECOND and two-stage PV-RCNN) and 3 benchmark 3D datasets (\textit{i.e.}, KITTI, nuScenes and Waymo). Extensive experiments evidence that the proposed Open-CRB demonstrates superiority and flexibility in recognizing both novel and known classes with very limited labeling costs, compared to state-of-the-art baselines. Source code is available at \url{https://github.com/Luoyadan/CRB-active-3Ddet/tree/Open-CRB}.
\end{abstract}


\section{Introduction}
\label{intro}  \IEEEPARstart{L}{iDAR} based 3D object detection is essential for understanding complex 3D scenes in various fields, including autonomous driving \citep{DBLP:conf/eccv/WangLGD20, DBLP:conf/nips/DengQNFZA21, DBLP:journals/pr/QianLL22a, chen2023redb} and robotics \citep{DBLP:conf/iros/AhmedTCMW18, DBLP:journals/sensors/WangLSLZSQT19, DBLP:conf/iros/MontesLCD20}. However, the success of these 3D models relies heavily on extensive training with substantial volumes of labeled 3D bounding boxes that have been manually labeled by human annotators. Accurately labeling a single 3D bounding box requires specifying seven degrees of freedom (DOF) — including position, size, and orientation — and can take over 100 seconds per annotation \citep{DBLP:conf/cvpr/SongLX15}. When a significant volume of fresh data arrives, manually labeling 3D boxes becomes both time-consuming and expensive.  To reduce the annotation burden, Active Learning (AL) proves valuable by selectively querying labels for a small fraction from a large pool of unlabeled data. The objective of AL selection criterion is to quantify the sample informativeness, using the heuristics derived from \textit{sample uncertainty} \citep{DBLP:conf/icml/GalIG17, DBLP:conf/iccv/DuZCC0021, DBLP:conf/cvpr/CaramalauBK21, DBLP:conf/cvpr/YuanWFLXJY21, DBLP:conf/iccv/ChoiELFA21, DBLP:conf/cvpr/Zhang0YWZH20, DBLP:conf/ijcai/ShiL19} and \textit{sample diversity} \citep{DBLP:conf/iclr/MaZMS21, DBLP:conf/cvpr/GudovskiyHYT20, DBLP:conf/eccv/GaoZYADP20, DBLP:conf/iccv/SinhaED19, DBLP:conf/nips/Pinsler0NH19}. These methods aim to optimize the model via learning from hard or diversely distributed samples. Recent research \citep{DBLP:conf/iclr/LuoCWYHB23} extends AL to LiDAR-based 3D object detection, employing a hierarchical active sampling strategy to acquire point clouds with concise labels, representative features, and geometric balance.




However, the design of existing AL algorithms is generally based on the \textit{closed world} assumption that the test data shares the \textbf{same} class set as the training data. This assumption does not always hold true in practical deployments of 3D object detectors, as real-world environments potentially include novel/unknown/out-of-distribution categories, referred to as \textit{open world} scenarios. To explore how to generalize 3D detectors to the practical open world scenario with minimal annotation costs, we introduce a new problem setting: Open World Active Learning for 3D Object Detection (OWAL-3D). Essentially different from traditional AL, OWAL-3D aims to capture a full spectrum of concepts within point clouds, including a sufficient number of instances from previously unknown classes. Human annotators then assign ground truth 3D bounding boxes and category labels (\textit{i.e.}, both known class and new class) for these instances. Optimizing the 3D detection model on this strategically selected subset allows the model to acquire knowledge of new concepts, thus effectively deployed in open world environments.

To seek solutions to OWAL-3D, we begin with preliminary experiments to assess whether existing AL policies and out-of-distribution (OOD) detection methods  \citep{DBLP:conf/iclr/SenerS18, DBLP:conf/iclr/MingSD023, liu2020energy_ood, DBLP:conf/nips/HuangGL21} can be directly applied to acquire point clouds which potentially contain unknown labels. We select the top 200 point clouds from the KITTI dataset \citep{DBLP:conf/cvpr/GeigerLU12} based on the highest scores determined by these methods. The selected point clouds are then assigned ground truth labels for all classes, including those novel ones not seen during pre-training, and are subsequently used to train the 3D detector for 30 epochs. The empirical results, illustrated in the bar plot of Figure \ref{fig:OLC}, reveal that diversity-based sampling methods, such as Coreset \citep{DBLP:conf/iclr/SenerS18} and Cider \citep{DBLP:conf/iclr/MingSD023}, tend to select point clouds with a large number of known labels. These approaches not only neglect unknown labels but also significantly increase annotation costs, undermining their practicality in the OWAL-3D setting. Conversely, uncertainty-based methods, such as ReAct \citep{liu2020energy_ood} and GradNorm \citep{DBLP:conf/nips/HuangGL21}, achieve a more balanced selection between unknown and known classes, meanwhile, achieving comparable results to those diversity-based methods. However, predicted 3D boxes with high uncertainty often contain low-quality objects, such as incomplete shapes or sparse points. Training on challenging point clouds may diminish the model's discriminative ability \citep{eskandar2024empirical}, and the limited performance of uncertainty-based methods in closed world AL for 3D detection also validate this \cite{DBLP:conf/iclr/LuoCWYHB23}.

\begin{figure*}[!t]
\centering
\includegraphics[width=1\linewidth]{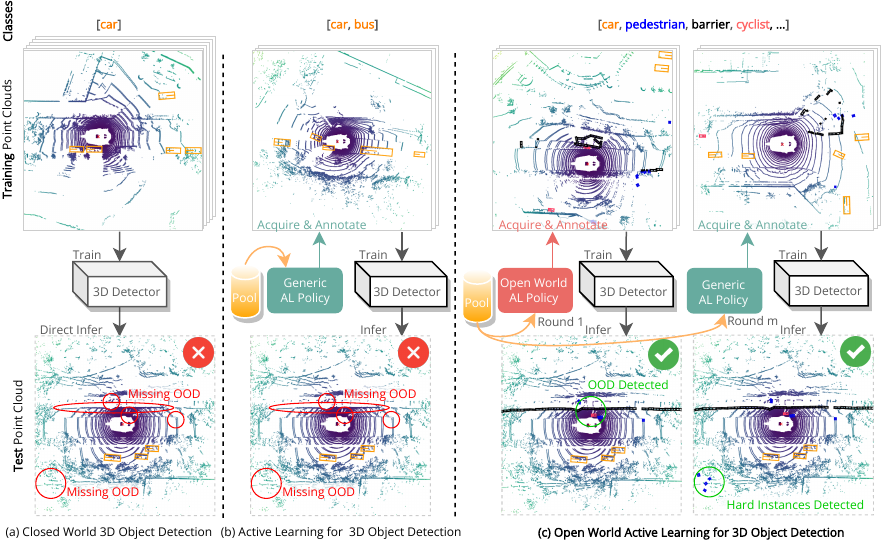}\vspace{-4ex}
\caption{\label{fig:owa_setting} The illustration of the Open World Active Learning for 3D Object Detection (OWAL-3D) and conventional tasks. In traditional closed world 3D detection (a), pre-trained 3D detectors struggle to localize and recognize objects from new classes (\textit{i.e.}, out-of-distribution (OOD)) in an open world context. Generic active learning (b) focuses on known categories, failing to select point clouds that potentially contain OODs. To address this, we introduce OWAL-3D (c), a framework that selectively acquires and labels a small subset of point clouds which are more likely to contain novel concepts using an Open World Active Learning (AL) policy. This approach enables the 3D detection model to efficiently generalize to new scenes containing novel object categories while significantly reducing time and cost.}
\end{figure*}

To tackle these challenges in OWAL-3D scenario, we propose a straightforward yet effective selection strategy: Open Label Conciseness (OLC), tailed for acquiring informative point clouds that are likely to contain novel labels while ensuring the quality of known labels. Specifically, the proposed OLC estimates the likelihood of unknown object labels existing in each point cloud by aggregating the uncertainty across all predicted bounding boxes. The OLC score is then calculated as the entropy of the predicted label distributions, including the estimated unknown labels, within each point cloud. Mathematically, the OLC policy can be interpreted into two relationships: (1) harmonic relationship of confidences across different classes, (2) inverse relationship between the number of boxes and their respective prediction confidences. The harmonic relationship ensures a high-quality set of object categories (\textit{i.e.}, diverse and concise) within selected point clouds. The inverse relationship aims to either seek \textit{more} 3D boxes with \textit{lower} confidence (for the discovery of novel concepts), or \textit{less} 3D boxes with \textit{higher} confidence (for reducing costs and keeping high-quality known labels). These two relationships, derived through OLC, strike an equilibrium between exploiting instances of novel classes and reducing annotation costs, while maintaining reliable known knowledge.


Based on our empirical results, the initial round of active learning using OLC uncovers a significant number of previously unknown labels, allowing the model to swiftly grasp the knowledge of new classes. Thus, any generic AL strategy can simply leverage OLC policy in the initial selection round to handle open world scenarios, making OLC a plug-and-play module, as illustrated in Figure \ref{fig:OLC}. Our final framework, Open-CRB, seamlessly integrates CRB with OLC by reverting to the CRB strategy for point cloud acquisition for the subsequent AL rounds.

A preliminary version of this work was presented in \citep{DBLP:conf/iclr/LuoCWYHB23}. We summarize the additional work and key contributions in this paper, as follows:
\begin{enumerate}
  \item We introduce a novel and more realistic task setting: Open World Active Learning for 3D Object Detection (OWAL-3D), which aims to efficiently generalize 3D detection models to open world environments that potentially contain unknown object classes. 
  \item To tackle OWAL-3D, we propose a simple yet effective plug-and-play open world AL policy, Open Label Conciseness (OLC) for discovery point clouds with novel labels and high-quality known labels and at minimal costs. 
  \item We develop a large-scale, open-source codebase for both open world and closed world AL in 3D object detection, supporting 15 baseline methods and 3 benchmark datasets to facilitate reproducibility and further research in this domain. We conduct extensive experiments with this codebase and our framework, Open-CRB, integrating OLC and CRB, demonstrates a 12.1\% improvement in mAP on the nuScenes dataset with only 50k annotated 3D boxes, compared to the best-performing baseline. The codebase is publicly available at \url{https://github.com/Luoyadan/CRB-active-3Ddet/tree/Open-CRB}.
\end{enumerate}

\section{Related Work}\label{sec2}

\subsection{Active Learning for Object Detection}

For a comprehensive review of classic active learning methods and their applications, we refer readers to \citep{10.1145/3472291}. Most active learning approaches were tailored for the image classification task, where the \textit{uncertainty} \citep{DBLP:conf/ijcnn/WangS14, DBLP:conf/icml/LewisC94,5206627margin, roth2006margin, Parvaneh_2022_CVPR_feature_mix, DBLP:conf/iccv/DuZCC0021, DBLP:conf/nips/KimSJM21, DBLP:conf/bmvc/BhatnagarGTS21} and \textit{diversity} \citep{DBLP:conf/iclr/SenerS18, DBLP:conf/iccv/ElhamifarSYS13, DBLP:conf/nips/Guo10, DBLP:journals/ijcv/YangMNCH15, DBLP:conf/icml/NguyenS04, DBLP:conf/iccv/0003R15, DBLP:conf/cvpr/AodhaCKB14} of samples are measured as the acquisition criteria. The hybrid works \citep{DBLP:conf/cvpr/KimPKC21, DBLP:conf/nips/CitovskyDGKRRK21, DBLP:conf/iclr/AshZK0A20, DBLP:journals/neco/MacKay92b, DBLP:conf/iccv/LiuDZLDH21, DBLP:conf/nips/KirschAG19, DBLP:journals/corr/abs-1112-5745} combine both paradigms such as by measuring uncertainty as to the gradient magnitude \citep{DBLP:conf/iclr/AshZK0A20} at the final layer of neural networks and selecting gradients that span a diverse set of directions. In addition to the above two mainstream methods, \citep{DBLP:conf/nips/SettlesCR07, roy2001toward-optimal-active, freytag2014selecting, DBLP:conf/cvpr/YooK19} estimate the expected model changes or predicted losses as the sample importance.

Lately, the attention of AL has shifted from image classification to the task of object detection \citep{DBLP:conf/cvpr/SiddiquiVN20,DBLP:conf/miccai/LiY20}. Early work \citep{DBLP:conf/bmvc/RoyUN18} exploits the detection inconsistency of outputs among different convolution layers and leverages the query by committee approach to select informative samples. Concurrent work \citep{DBLP:conf/accv/KaoLS018} introduces the notion of localization tightness as the regression uncertainty, which is calculated by the overlapping area between region proposals and the final predictions of bounding boxes. Other uncertainty-based methods attempt to aggregate pixel-level scores for each image \citep{Aghdam_2019_ICCV}, reformulate detectors by adding Bayesian inference to estimate the uncertainty \citep{DBLP:conf/icra/HarakehSW20} or replace conventional detection head with the Gaussian mixture model to compute aleatoric and epistemic uncertainty \citep{DBLP:conf/iccv/ChoiELFA21}. {\color{black} A hybrid method \citep{DBLP:conf/cvpr/WuC022} considers image-level uncertainty calculated by entropy and instance-level diversity measured by the similarity to the prototypes. Lately, the AL technique has been leveraged for transfer learning by selecting a few uncertain labeled source bounding boxes with high transferability to the target domain, where the transferability is defined by domain discriminators  \citep{9548667,DBLP:journals/tci/Al-SaffarBBTGA21}. Inspired by neural architecture searching, \cite{DBLP:conf/bmvc/TangJWXZ0LX21} adopted the ‘swap-expand’ strategy to seek a suitable neural architecture including depth, resolution, and receptive fields at each active selection round. Recently, some works augment the Weakly-Supervised Object Detection (WS-OD) with an active learning scheme. In WS-OD, only image-level category labels are available during training. Some conventional AL methods such as predicted probability, and probability margin are explored in \citep{wang2022weaklySupervisedObject}, while in \citep{vo2022activeLearningStrategies}, “box-in-box" is introduced to select images where two predicted boxes belong to the same category and the small one is “contained” in the larger one.} Nevertheless, it is not trivial to adapt all existing AL approaches for 2D detection as ensemble learning and network modification lead to more model parameters to learn, which could be hardly affordable for 3D tasks. 

Active learning for 3D object detection has been relatively under-explored than other tasks, potentially due to its large-scale nature. Most existing works \citep{DBLP:conf/ivs/FengWRMD19, DBLP:conf/ivs/SchmidtRTK20} simply apply the off-the-shelf generic AL strategies and use hand-crafted heuristics including Shannon entropy \citep{DBLP:conf/ijcnn/WangS14}, ensemble \citep{DBLP:conf/cvpr/BeluchGNK18}, localization tightness \citep{DBLP:conf/accv/KaoLS018}, {Mc-dropout} \citep{DBLP:conf/icml/GalG16} and neural tangent kernel \citep{luo2023kecor} for 3D detection learning. However, the abovementioned solutions are base on the cost of labeling point clouds rather than the number of 3D bounding boxes, which inherently are biased to the point clouds containing more objects. However, in our work, the proposed {CRB} greedily searches for the unique point clouds while maintaining the same marginal distribution for generalization, which implicitly queries objects to annotate without repetition and saves labeling costs.

\subsection{Open World Object Detection}

The Open World Object Detection (OWAD) task, introduced recently in the work of \cite{DBLP:conf/cvpr/Joseph0KB21}, has garnered considerable attention within the research community, owing to its potential real-world applications. \cite{DBLP:conf/cvpr/Joseph0KB21} propose an ORE approach that enhances the faster-RCNN model's ability to recognize and learn unknown objects, by feature-space contrastive clustering, an RPN-based unknown object detector, and an Energy-Based Unknown Identifier. Building upon the ORE, \cite{DBLP:conf/icip/YuML0X22} further extended the methodology by addressing the issue of distribution overlap between known and unknown classes in feature space embeddings, reducing the confusion that often arises when distinguishing between known and unknown objects. Simultaneously, \cite{DBLP:conf/mm/WuZMWL22} endeavored to extend ORE by introducing an additional objectiveness detection head that predicts the Intersection over Union (IoU) between the localized bounding boxes and the corresponding ground truth boxes. In an effort to refine the decision boundaries of known and unknown classes, \cite{DBLP:conf/cvpr/MaLZGZG023} proposes to decouple the known and unknown features, thus promoting both known and unknown object recognition.

In recent times, there has been a notable surge in the adaptation of transformer-based techniques in the context of Open World Object Detection (OW-OD). The pioneering work by \cite{DBLP:conf/cvpr/GuptaNJ0KS22} introduced OW-DETR, an adaptation of the Deformable DETR model tailored to confront the specific challenges posed by OW-OD tasks. OW-DETR leverages a pseudo-labeling approach to supervise the detection of unknown objects, wherein unmatched object proposals with strong backbone activations are characterized as potential unknown objects. Seeking to enhance the localization capabilities of transformer-based object detectors, \cite{DBLP:conf/eccv/MaazR0KA022} developed Multi-modal Vision Transformers (MViT) to align image-text pairs. Since textual language descriptions convey high-level information, the fusion of modalities aids in learning fairly generalizable properties of universal object categories. Different from MViT, which relies on language modality to improve the model, PROB \citep{DBLP:conf/cvpr/ZoharWY23} integrates probabilistic models into existing OW-DETR framework to facilitate objectiveness estimation within the embedded feature space. Furthermore, an evolution of the OW-DETR model is presented in the form of the LoCalization and IdentificAtion Cascade Detection Transformer (CAT) by Ma et al. \citep{DBLP:conf/cvpr/MaW0FLLL23}. CAT aims to emulate human thinking patterns, which inherently prioritizes the initial detection of all foreground objects before delving into detailed recognition. To achieve this, CAT decouples the detection process in the cascade decoding way to prioritize localization before classification, enhancing the model's capacity to identify and retrieve unknown objects in open world environments.

However, current OW-OD methods typically require a large amount of manually labeled data for learning each of the new tasks. In contrast, the OWAL-3D tackled in this paper significantly reduces costs, thus more efficiently gaining new concepts from the open world environments.

\section{Preliminaries}

\subsection{Problem Definition of CWAL-3D}
In this section, we mathematically formulate the task of Closed World Active Learning for 3D Object Detection (CWAL-3D) and set up the notations.

\begin{definition}[3D Object Detection] Given an orderless LiDAR point cloud $\mathcal{P} = \{x, y, z, e\}$ with 3D location $(x, y, z)$ and reflectance $e$, the goal of 3D object detection is to localize the objects of interest as a set of 3D bounding boxes $\mathcal{B} = \{b_k\}_{k\in[N_B]}$ with $N_B$ indicating the number of detected bounding boxes, and predict the associated box labels $Y = \{y_k\}_{k\in[N_B]} \in\mathcal{Y} = \{1,\ldots,C\}$, with $C$ being the number of classes to predict. 

Each bounding box $b$ represents the relative center position $(p_x, p_y, p_z)$ to the object ground planes, the box size $(l, w, h)$, and the heading angle $\theta$. Mainstream 3D object detectors \textcolor{black}{use point clouds $\mathcal{P}$ to extract point-level features $\bm{x}\in\mathbb{R}^{W\cdot L\cdot  F}$ ~}\citep{DBLP:conf/cvpr/ShiWL19,DBLP:conf/iccv/YangS0SJ19,DBLP:conf/cvpr/YangS0J20} or by voxelization \citep{DBLP:conf/cvpr/ShiGJ0SWL20}, with $W$, $L$, $F$ representing width, length, and channels of the feature map. The feature map $\bm{x}$ is passed to a classifier $f(\cdot; \bm{w}_{f})$ parameterized by $\bm{w}_{f}$ and regression heads $g(\cdot; \bm{w}_{g})$ (\textit{e.g.,} box refinement and ROI regression) parameterized by $\bm{w}_{g}$. The output of the model is the detected bounding boxes $\widehat{\mathcal{B}} = \{\hat{b}_k\}$ with the associated box labels $\widehat{Y}=\{\hat{y}_k\}$ from anchored areas. \textcolor{black}{The loss functions $\ell^{cls}$ and $\ell^{reg}$ for classification (\textit{e.g.}, regularized cross entropy loss \cite{DBLP:journals/corr/abs-1808-09540}) and regression (\textit{e.g.}, mean absolute error/$L_1$ regularization \cite{DBLP:journals/spl/QiDSML20}) are assumed to be Lipschitz continuous.} 
\end{definition}

\begin{definition}[CWAL-3D]
In an active learning pipeline, a small set of labeled point clouds $\mathcal{D}_L=\{(\mathcal{P}, \mathcal{B}, Y)_i\}_{i\in[m]}$ and a large pool of raw point clouds $\mathcal{D}_U=\{(\mathcal{P})_j\}_{j\in[n]}$ are provided at training time, with $n$ and $m$ being a total number of point clouds and $m\ll n$. For each active learning round $r\in[R]$, and based on the criterion defined by an active learning policy, we select a subset of raw data $\{\mathcal{P}_j\}_{j\in[N_r]}$ from $\mathcal{D}_U$ and query the labels of 3D bounding boxes from an oracle $\bm{\Omega}: \mathcal{P}\rightarrow \mathcal{B}\times\mathcal{Y}$ to construct $\mathcal{D}_S=\{(\mathcal{P}, \mathcal{B}, Y)_j\}_{j\in[N_r]}$. The 3D detection model is pre-trained with $\mathcal{D}_L$ for active selection, and then retrained with $\mathcal{D}_{S}\cup \mathcal{D}_L$ until the selected samples reach the final budget $B$, \textit{i.e.,} $\sum_{r=1}^{R}N_{r} = B$.     
\end{definition}

\subsection{CWAL-3D: CRB Approach}

The CRB framework, proposed in our preliminary work \citep{DBLP:conf/iclr/LuoCWYHB23}, differs from generic active learning (AL) policies by being specifically designed for 3D object detection. It achieves this by acquiring label-\textbf{C}oncise, feature-\textbf{R}epresentative, and geometrically \textbf{B}alanced point clouds while minimizing annotation costs. The framework employs a hierarchical filtering process to select samples that meet the three specific criteria above. First, we choose $\mathcal{K}_1$ candidates through label-concise sampling to avoid redundancy within the point cloud. Recognizing the equal importance of object category classification and 3D box regression in this task, we then select $\mathcal{K}_2$ representative prototypes, with $\mathcal{K}_1, \mathcal{K}_2 \ll n$. This selection process incorporates both 3D box classification and regression to ensure that the prototypes contain representative object features. Finally, a greedy search is used to identify $N_r$ prototypes that align with the prior marginal distribution of the test data, ensuring that the geometric characteristics of the selected 3D boxes are balancedly distributed. This hierarchical sampling approach reduces the cost by $\mathcal{O}((n-\mathcal{K}_1)T_2 + (n-\mathcal{K}_2)T_3)$, where $T_2$ and $T_3$ denote the runtime of criterion evaluation. We present a detailed explanation of each selection criterion, along with the theoretical guarantees, in the original paper \citep{DBLP:conf/iclr/LuoCWYHB23}. While CRB demonstrates significant performance gain over existing AL methods for CWAL-3D, its effectiveness, in an open world scenario with potential novel categories, remains to be explored.

\section{Our Approach: Open-CRB}
\subsection{Problem Definition of \textcolor{red}{O}WAL-3D}
\begin{definition}[OWAL-3D]
In the open world scenarios, unlabeled pool $\mathcal{D}_O$ usually contain $U$ \textbf{\textit{novel}} / \textbf{\textit{unknown}} classes $\{C+1, \ldots,C+U\}$ which do not exist in $\mathcal{D}_L$, while the off-the-shelf 3D object detector is pretrained on a limited set $\mathcal{D}_L$. Different from CWAL-3D, the objective of OWAL-3D is to maximize the 3D detection performance on all classes (\textit{i.e.}, both known and unknown) by training the detector on the subset $\{\mathcal{P}_j\}_{j\in[N_r]}$ strategically selected from $\mathcal{D}_O$.
\end{definition}


\begin{figure*}[!t]
\centering
\includegraphics[width=0.95\linewidth]{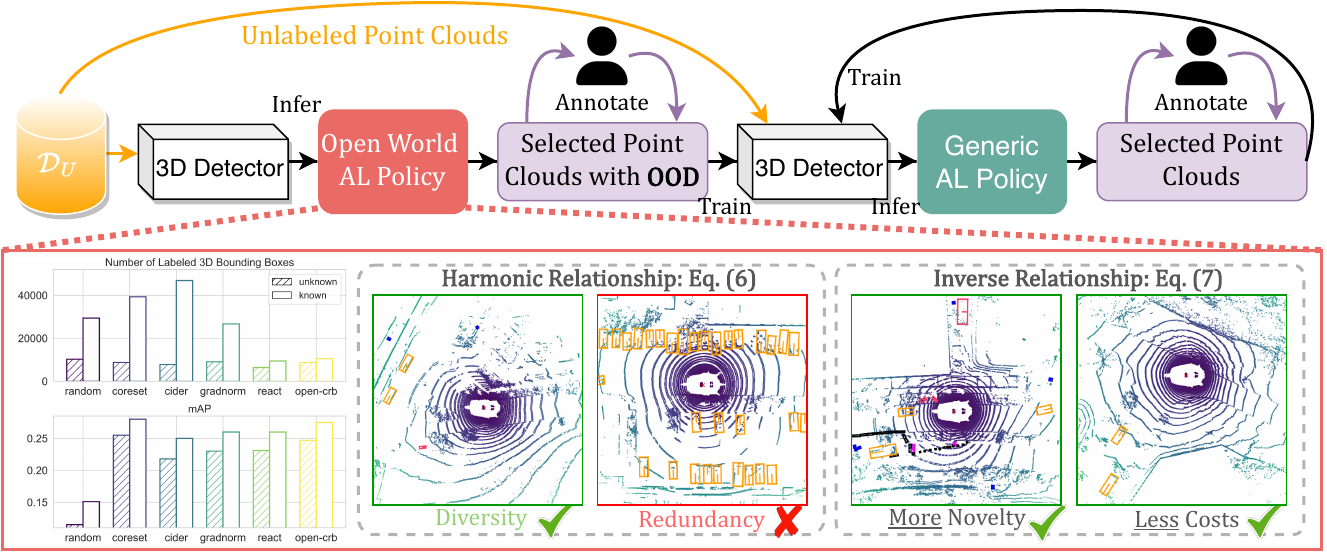}\vspace{-1ex}
\caption{\textbf{Upper}: The overall framework of the proposed Open-CRB for OWAL-3D. \textbf{Lower}: The illustration of the proposed open world AL policy, Open Label Conciseness (OLC), which is designed for active selection from an unlabeled open world pool.  The \textit{left} bar plots report the annotation costs of the baseline methods and the proposed Open-CRB in the first selection round, along with the detection performance after training on the selected point clouds. The visualized point clouds in the \textit{middle} and \textit{right} illustrate the selection criteria (Eq.  \eqref{equ:final_entropy}), guided by two key relationships (Remark \ref{rmk:3}). The first relationship ensures a harmonic balance among the confidences associated with different predicted classes, promoting diversity and minimizing redundancy within the selected point clouds. The second relationship is inversely proportional, linking the number of bounding boxes to confidence levels. This relationship either 1) encourages exploration of unknown objects when low-confidence predictions are abundant, or 2) reduces the number of bounding boxes when the likelihood of unknown objects is low. These dual relationships work in tandem to select point clouds that include concise and high-quality known labels, and more unknown labels. The detailed algorithm is clearly summarized in Algorithm \ref{alg:open-crb}.}
\label{fig:OLC}
\end{figure*}

\noindent \textbf{Discussion: from closed world AL to open world AL.} To explore whether existing Active Learning (AL) and Out-of-Distribution (OOD) methods can acquire new knowledge from open world data, we conducted a pilot study. As shown in the \textit{bar plots} of Figure \ref{fig:OLC}, it is evident that diversity-based methods tend to select a large number of redundant known 3D boxes (\textit{i.e.}, Coreset selected 39,350 boxes, and Cider chose 46,847). In contrast, methods selecting point clouds with high prediction uncertainty often lead to fewer and harder objects. For example, GradNorm selected 26,716 known boxes, and ReAct chose only 9,440. It can also be observed that both diversity-based and uncertainty-based methods acquire approximately the same number of unknown/novel instances, around 10,000. 

Despite the fact that annotation costs of uncertainty-based methods are significantly reduced, comparable performance is still maintained. As illustrated in the lower bar plot of Figure \ref{fig:OLC}, ReAct achieved an unknown mAP of 0.23 with a total cost of only 15,848 labeled boxes, whereas Coreset, despite incurring 48,133 labeled boxes (204\% higher cost), demonstrated only a marginal mAP improvement of 11.2\%. This finding validates our core idea that maintaining a well-balanced ratio between unknown and known samples leads to satisfactory detection performance with minimal overall cost. For example, in the case of ReAct, the ratio is 0.68, while for diversity-based methods, such as Cider, the ratio is much lower which is 0.17. Motivated by this, our proposed Open Label Conciseness (OLC) ensures that the selected point clouds likely contain high-uncertainty instances while maintaining diverse, concise, and high-quality known objects. The experiment demonstrates that in the first round of selection, OLC selected only 19,232 labeled boxes with a high unknown-to-known ratio of 0.83, achieving strong performance with an mAP of 0.28. Since OLC is able to acquire sufficient new knowledge from the open world in the first selection round, subsequent selection rounds can seamlessly integrate with any generic method, as shown in Table \ref{tab:abl_crb_vs_open_crb}. For instance, after using OLC in the first round, switching to GradNorm or Cider in the following rounds achieves improvements of 35.73\% and 10.1\%, respectively, compared to using only GradNorm or Cider throughout. This demonstrates that the proposed OLC module is plug-and-play, capable of transforming an open world problem into a closed-world one with only a single round of active selection, thereby effectively supporting any closed-world generic strategy.

\subsection{Open Label Conciseness (OLC) Sampling}


We begin by introducing the estimation of the unknown label component for each point cloud. A straightforward approach is applied to sum the uncertainty across all predicted boxes. Accordingly, the estimated unknown component for the $j$-th point cloud, treated as an additional class $C+1$, is formulated as:
\begin{align} \label{equ:unk_component}
   \quad \bm{p}_{j,C+1} &= \frac{\sum_{i=1}^{N_B} (1-\tilde{y}_i)}{N_B},
\end{align}
where $\tilde{y}_i$ represents the confidence of the $i$-th predicted box, and $N_B$ is the total number of box predictions. We then formulate the known label component for each class, based on the prediction confidence, as follows: 
\begin{equation}\label{equ:known_component}
\begin{aligned}
   \bm{p}_{j,c} = \frac{\sum_{i=1}^{N_B} \mathbbm{1}(\hat{y}_i = c) \times \tilde{y}_i}{N_B},~\text{for}~ c\in[1, \cdots, C].
\end{aligned}
\end{equation}

\noindent Leveraging Eq. \eqref{equ:unk_component} and Eq. \eqref{equ:known_component}, the OLC score for each point cloud is estimated by calculating the unknown-aware entropy of the label distribution, as follows:
\begin{align} \label{equ:final_entropy}
  \quad \tilde{H}(\widehat{Y}_{j, S}) &= -\sum_{c=1}^{C+1}\bm{p}_{i,c}\log \bm{p}_{i,c}.
\end{align}

\noindent We compute the OLC score for all point clouds in the unlabeled pool, then select the top K point clouds with the highest scores for manual labeling and use them to train the 3D detection model.

\begin{remark}\label{rmk:3}
To discuss the properties of the derived OLC criterion, we give a simple example as below. Given that 3D object detector was pre-trained on \textbf{two} known classes denoted as $1$ and $2$, there exists a new class in the unlabeled pool. When testing on an arbitrary point cloud $j$ sourced from the open world, we have $n_1$ and $n_2$ box predictions of class $1$ and $2$, respectively. Note that $n_1 + n_2 = N_B$. The average prediction confidence for classes $1$ and $2$ are $\bar{p}_1=\frac{\bm{p}_{j,1}}{n_1}N_B$ and  $\bar{p}_2=\frac{\bm{p}_{j,2}}{n_2}N_B$. Referring to Eq. \eqref{equ:unk_component}, the unknown label components can be simplified: 
\begin{align}
    &\bm p_{j, C+1} N_B = n_1(1-\bar{p}_1)+n_2(1-\bar{p}_2),\\
    &N_B- \sum_{i}^{N_B} \tilde{y}_i = (n_1 + n_2) - \sum_{i}^{N_B} \tilde{y}_i \nonumber
\end{align}
The most desired case is when label entropy $\tilde{H}(\widehat{Y}_{j, S})$ is maximized, we have, 
\begin{equation}\label{equ:4eq}
    n_1\bar{p}_1=n_2\bar{p}_2=n_1(1-\bar{p}_1)+n_2(1-\bar{p}_2).
\end{equation}It can be interpreted as achieving maximum diversity in selecting both known and unknown classes equally. This equation leads to the following two relationships:

\noindent \textbf{Harmonic Relationship:} By substituting $n_1$ and $n_2$ in the Eq. \eqref{equ:4eq}, we can obtain the harmonic relationship between the averaged known confidences $\bar{p}_1$ and $\bar{p}_2$:
\begin{align} \label{eq:harmonic}
\quad\frac{2\bar{p}_1\bar{p}_2}{\bar{p}_1+\bar{p}_2}=const.
\end{align}
The constant equals to $2/3$ when $C$=2. This relationship is a harmonic mean of the averaged confidence among the known classes. If a class is absent, this relationship will be difficult to maintain. Hence, this relation guarantees that the selected point cloud contains a diverse category distribution. As shown in the left two scenarios of Figure \ref{fig:OLC}, the left example is preferred as it contains objects of multiple different categories, while the right one will lead to very low label entropy and not be selected by OLC.

\noindent \textbf{Inverse Relationship:} On the other side, when we substitute $\bar{p}_1$ or $\bar{p}_2$ in Eq. \eqref{equ:4eq}, we can derive the following constraints between the averaged confidence and the selected instance numbers: 
\begin{align}\label{eq:inverse}
\quad\frac{n_2}{n_1} \:\propto\: \bar{p}_1, \quad\frac{n_1}{n_2} \:\propto\: \bar{p}_2.
\end{align}

This equation shows an inverse relationship between $n_1$ and $\bar{p}_1$, and for $n2$ and $\bar{p}_2$ vice versa. When $n_2$ is fixed and samples are of low confidence $\bar{p}_1 \downarrow$, this criterion will lead to picking more such instances ($n_1\uparrow$). This selection rule can help identify more unknown instances, as illustrated in the third example of Figure \ref{fig:OLC}. Conversely, A high averaged confidence $\bar{p}_1\uparrow$ generally indicates a high likelihood of the point cloud containing a familiar known class, thus this equation will penalize $n_1\downarrow$ to minimize the number of boxes, as depicted in the last case of Figure \ref{fig:OLC}. 

Therefore, the harmonic relationship compels the AL strategy to favor the point clouds with various categories, whereas the inverse relationship dynamically mines objects of novel categories or preserves annotations. These dual relationships constrain each other to ensure the selection of point clouds not only with a diverse and concise set of classes in limited numbers, but also with a high likelihood to contain unknown categories.

\end{remark}

\begin{algorithm}[t]
\caption{Open-CRB for 3D Object Detection}\label{alg:open-crb}
\begin{algorithmic}
\State Pre-train the 3D detector using the initial set of point clouds with closed set labels until convergence. 
\While{budget allows} 
\If{selecting from open world pool} 
\State Sample point clouds with top OLC scores (Eq. \eqref{equ:final_entropy}).
\EndIf
\If{selecting from closed world pool} 
\State Shift to CRB \cite{DBLP:conf/iclr/LuoCWYHB23} for  point clouds acquisition.
\EndIf
\State Annotate the newly selected point clouds with the oracle, and then re-train the 3D detector.
\EndWhile
\end{algorithmic}
\end{algorithm}

\section{Experiments}

\subsection{Datasets}

\textbf{KITTI} \citep{DBLP:conf/cvpr/GeigerLU12} is one of the most representative datasets for point cloud-based object detection. The dataset consists of 3,712 training samples (\textit{i.e.,} point clouds) and 3,769 \textit{val} samples. The dataset includes a total of 80,256 labeled objects with three commonly used classes for autonomous driving: cars, pedestrians, and cyclists. To fairly evaluate baselines and the proposed method on KITTI dataset \citep{DBLP:conf/cvpr/GeigerLU12}, we follow the work of~\citep{DBLP:conf/cvpr/ShiGJ0SWL20}: we utilize Average Precision (AP) for 3D and bird eye view (BEV) detection, and the task difficulty is categorized to {Easy}, {Moderate}, and {Hard}, with a rotated IoU threshold of $0.7$ for cars and $0.5$ for pedestrian and cyclists. The results evaluated on the validation split are calculated with $40$ recall positions.

\noindent \textbf{Waymo Open} dataset \citep{DBLP:conf/cvpr/SunKDCPTGZCCVHN20} is a challenging testbed for autonomous driving comprised of high resolution sensor data, containing 158,361 training samples and 40,077 testing samples. The point clouds contain 64 lanes of LiDAR corresponding to 180k points every 0.1s. To evaluate on Waymo dataset \citep{DBLP:conf/cvpr/SunKDCPTGZCCVHN20}, we adopt the officially published evaluation tool for performance comparisons, which utilizes AP and the average precision weighted by heading (APH). The respective IoU thresholds for vehicles, pedestrians, and cyclists are set to 0.7, 0.5, and 0.5. Regarding detection difficulty, the Waymo test set is further divided into two levels. {Level 1} (and {Level 2}) indicates there are more than five inside points (at least one point) in the ground-truth objects. To alleviate computation overhead, we set the sampling interval to 10. 

\noindent \textbf{nuScenes} dataset \citep{DBLP:conf/cvpr/CaesarBLVLXKPBB20} comprises a total of 1000 driving sequences, which have been partitioned into three distinct subsets for the purposes of training, validation, and testing, encompassing 700, 150, and 150 sequences, respectively. These sequences, each possessing a temporal span of approximately 20 seconds, are characterized by a LiDAR data acquisition frequency of 20 frames per second (FPS). The nuScenes dataset \citep{DBLP:conf/cvpr/CaesarBLVLXKPBB20} adopts two metrics: mean average precision (mAP) and nuScenes detection score (NDS). The former one is commonly employed but in nuScenes, they leverage the 2D center distance within the ground plane rather than IoU-based affinities. The latter is a weighted metric based on mAP and average error of translation, scale, orientation, velocity, and attribute.

\subsubsection{Evaluation Metric for OWAL-3D}

To thoroughly assess the effectiveness of various methods within the OWAL-3D context, we establish three specific metrics tailored to open world active learning for 3D object detection. These metrics measure the methodology's capacity to 1) explore unknown classes, 2) accurately recognize all classes, and 3) save the associated annotation costs. 

\noindent \textbf{Performance across unknown classes}: $\text{mAP}_{unk}$. We average the AP score for all unknown categories: 
\begin{align}
   \quad \text{mAP}_{unk} = \frac{1}{U} \sum^{U}_{i=1} \text{AP}_i,  i \in \{1, \ldots, U\}
\end{align}
where $\text{AP}_i$ indicates the AP score of ($C+i$)-the class, calculated by KITTI or nuScenes official metric. A high score of $\text{mAP}_{unk}$ signifies that the 3D detection model has gained sufficient knowledge of the novel class within the open world.

\noindent \textbf{Balanced performance across all classes}: $\text{mAP}_{H}$. We compute the harmonic mean between $\text{mAP}_{unk}$ and the mean AP across known classes $\text{mAP}_{k}$ as:
\begin{align}
   \quad &\text{mAP}_{H} = \frac{2}{1/{\text{mAP}_{unk}}+1/{\text{mAP}_{k}}}, \\
    &\text{mAP}_{k} = \frac{1}{C} \sum^{C}_{j=1} \text{AP}_j, j \in \{1,\ldots,C\}.
\end{align}
The harmonic mean was widely adopted in previous work \citep{6365193, 9040673, 10107906} related to open-set tasks, which helps to prevent any bias towards either unknown classes or known classes, indicating balanced and unbiased results.

\begin{figure}[h]%
        \subfloat{{\includegraphics[width=0.245\textwidth]{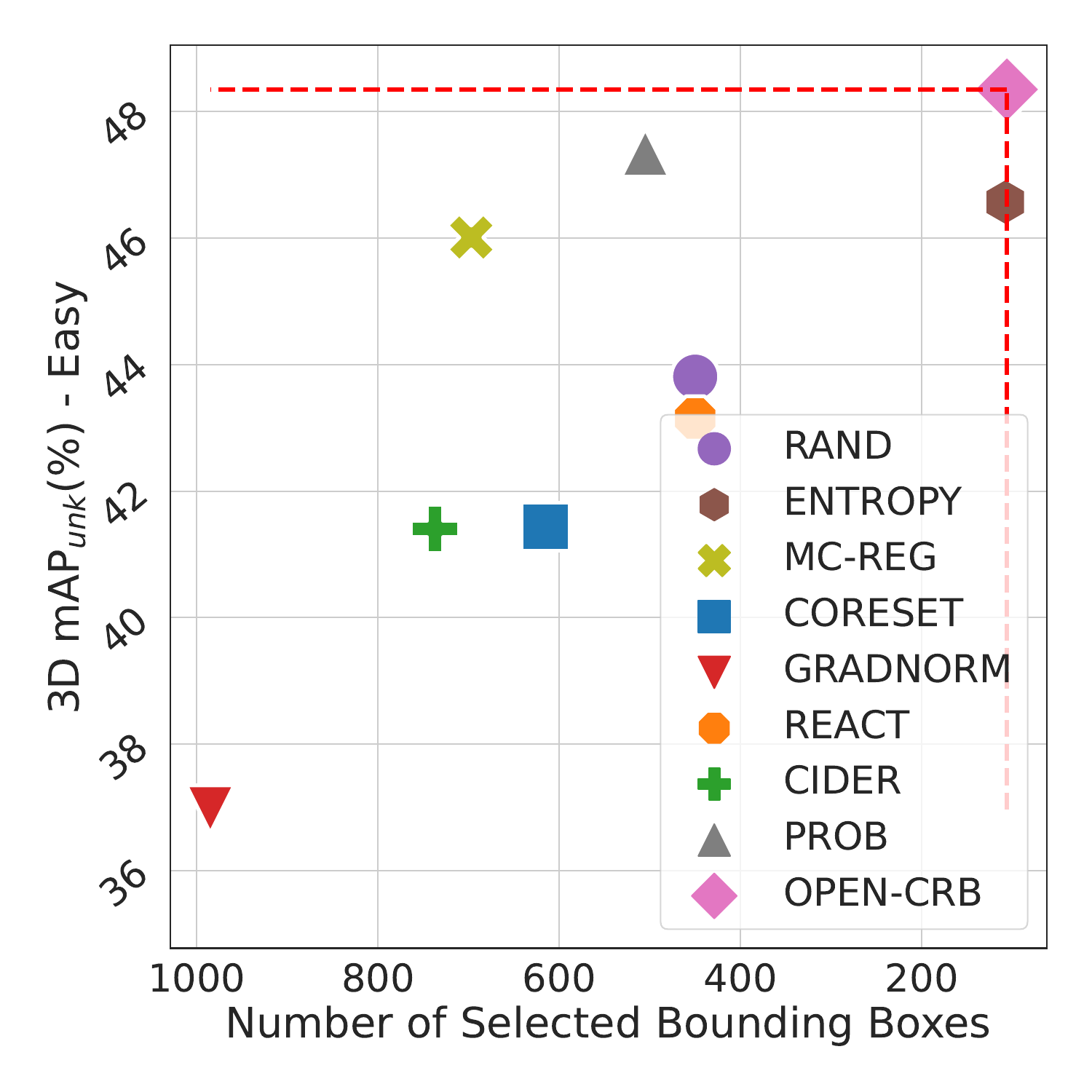} }}%
    \subfloat{{\includegraphics[width=0.245\textwidth]{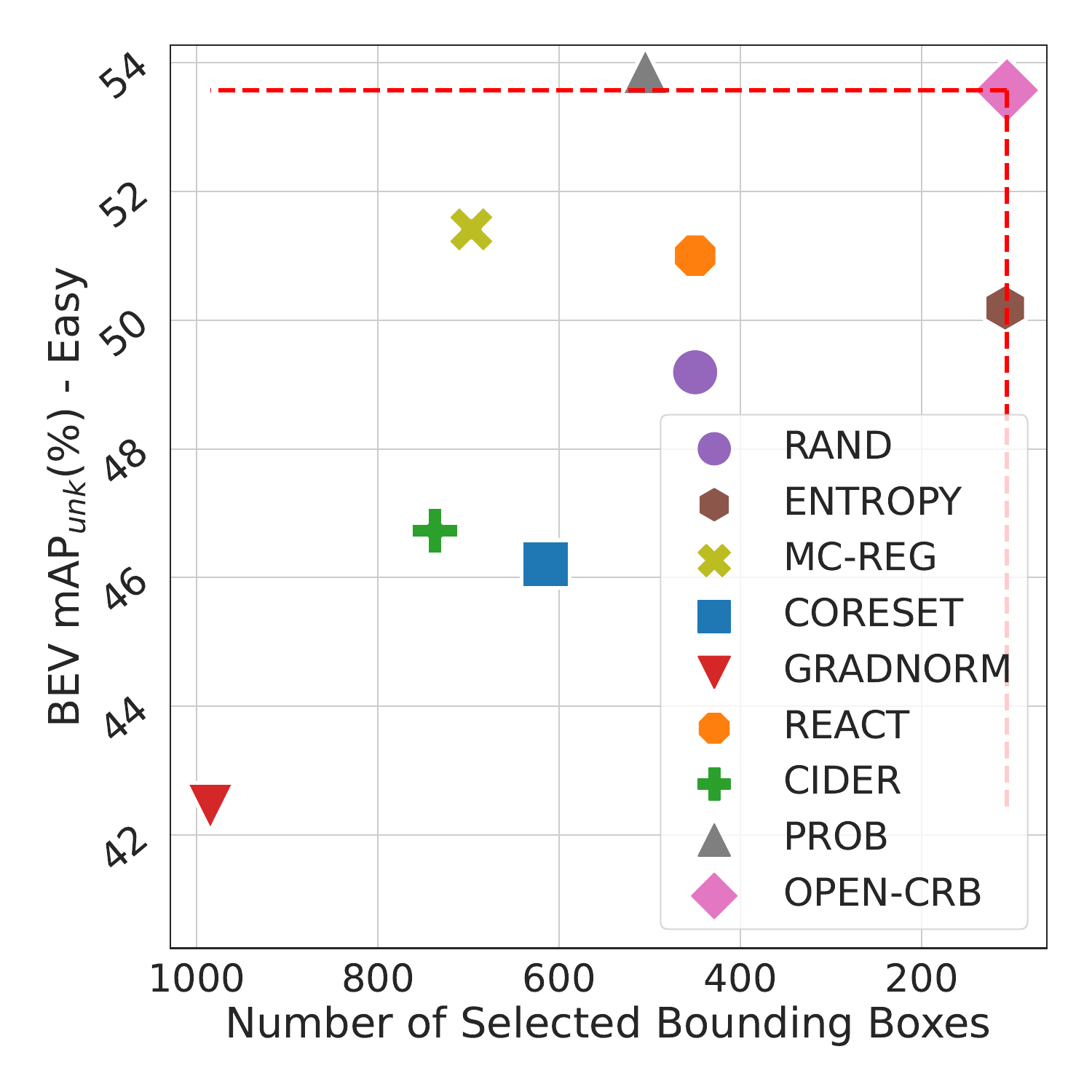} }}%
   \hfill
    \subfloat{{\includegraphics[width=0.245\textwidth]{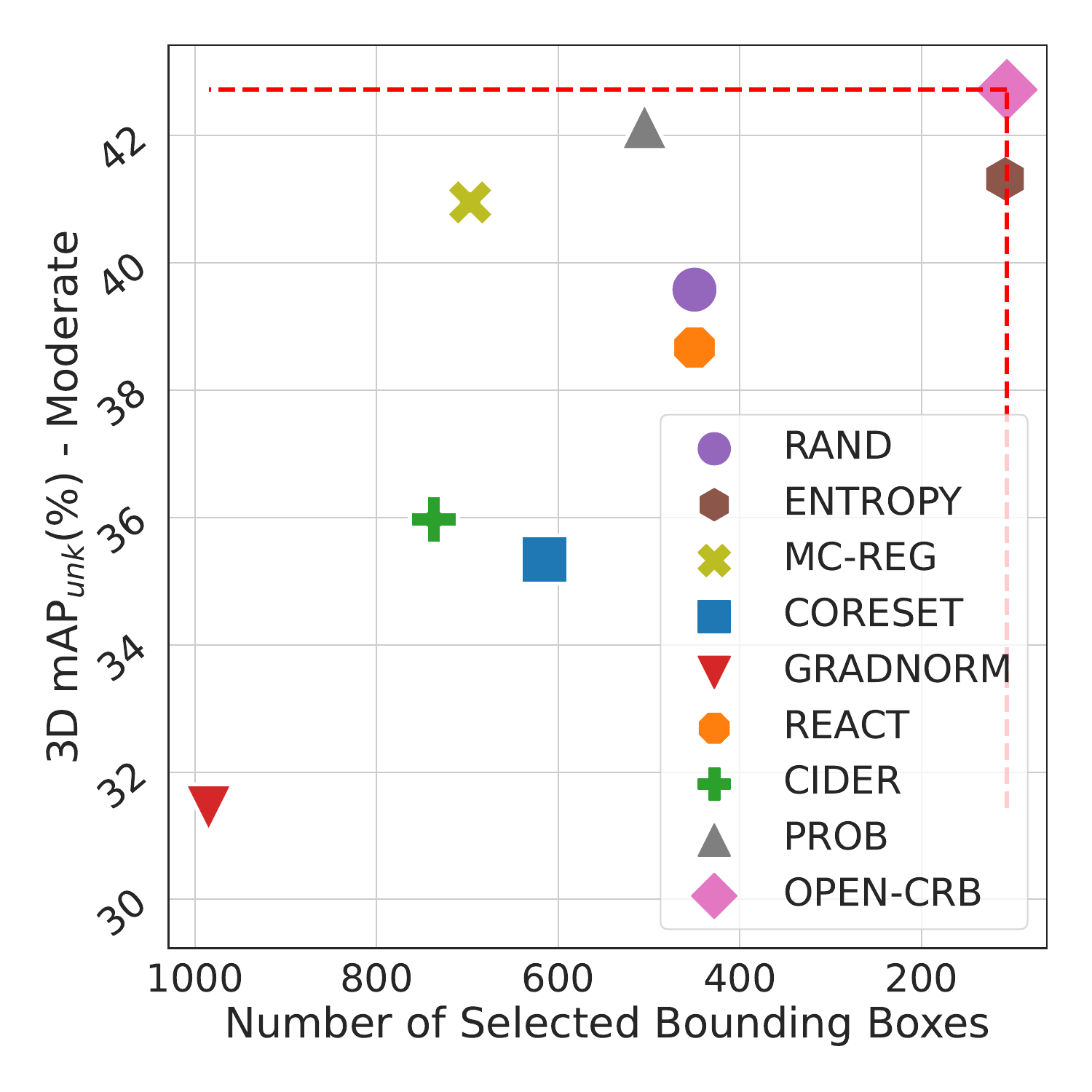} }}%
    \subfloat{{\includegraphics[width=0.245\textwidth]{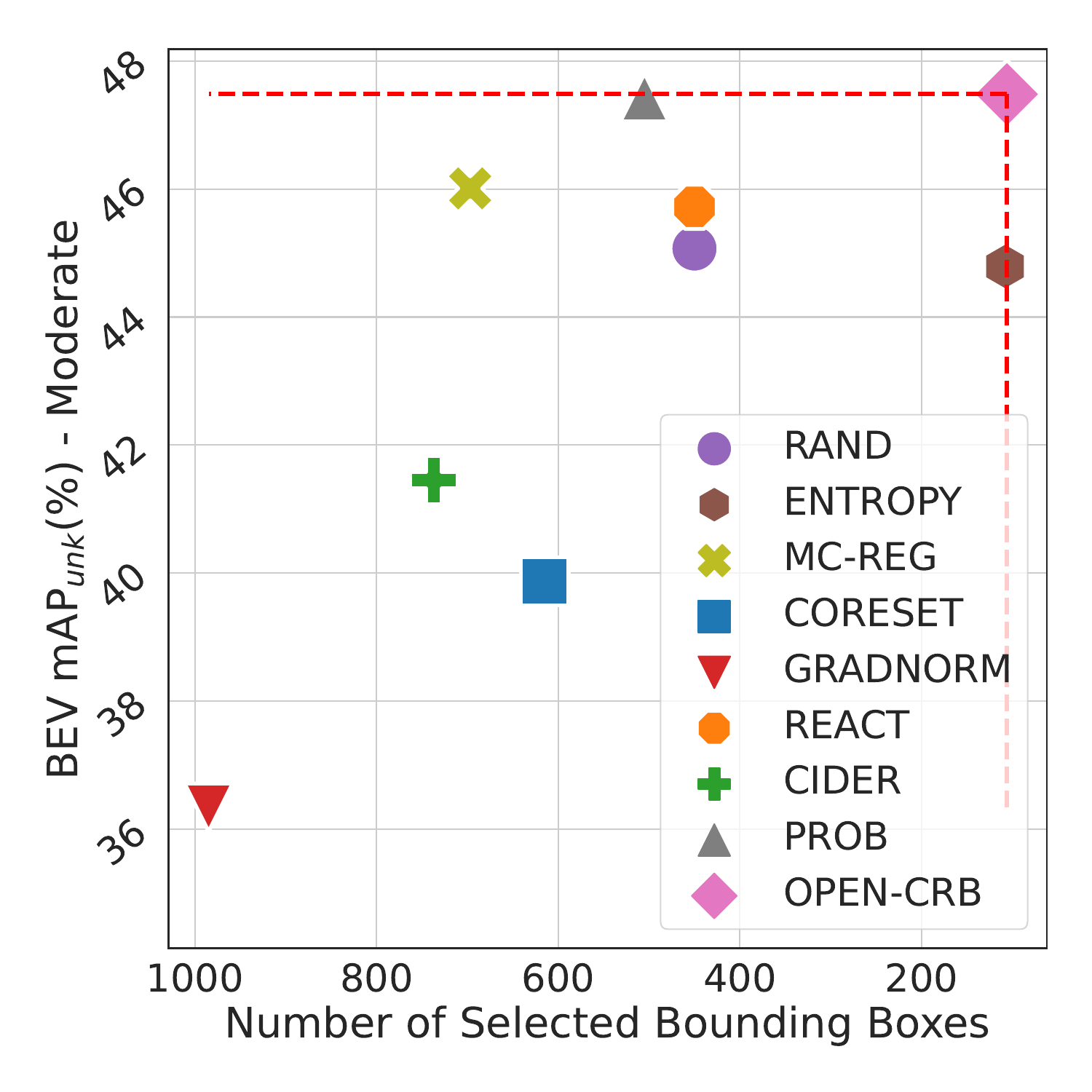} }}%
   \hfill
    \subfloat{{\includegraphics[width=0.245\textwidth]{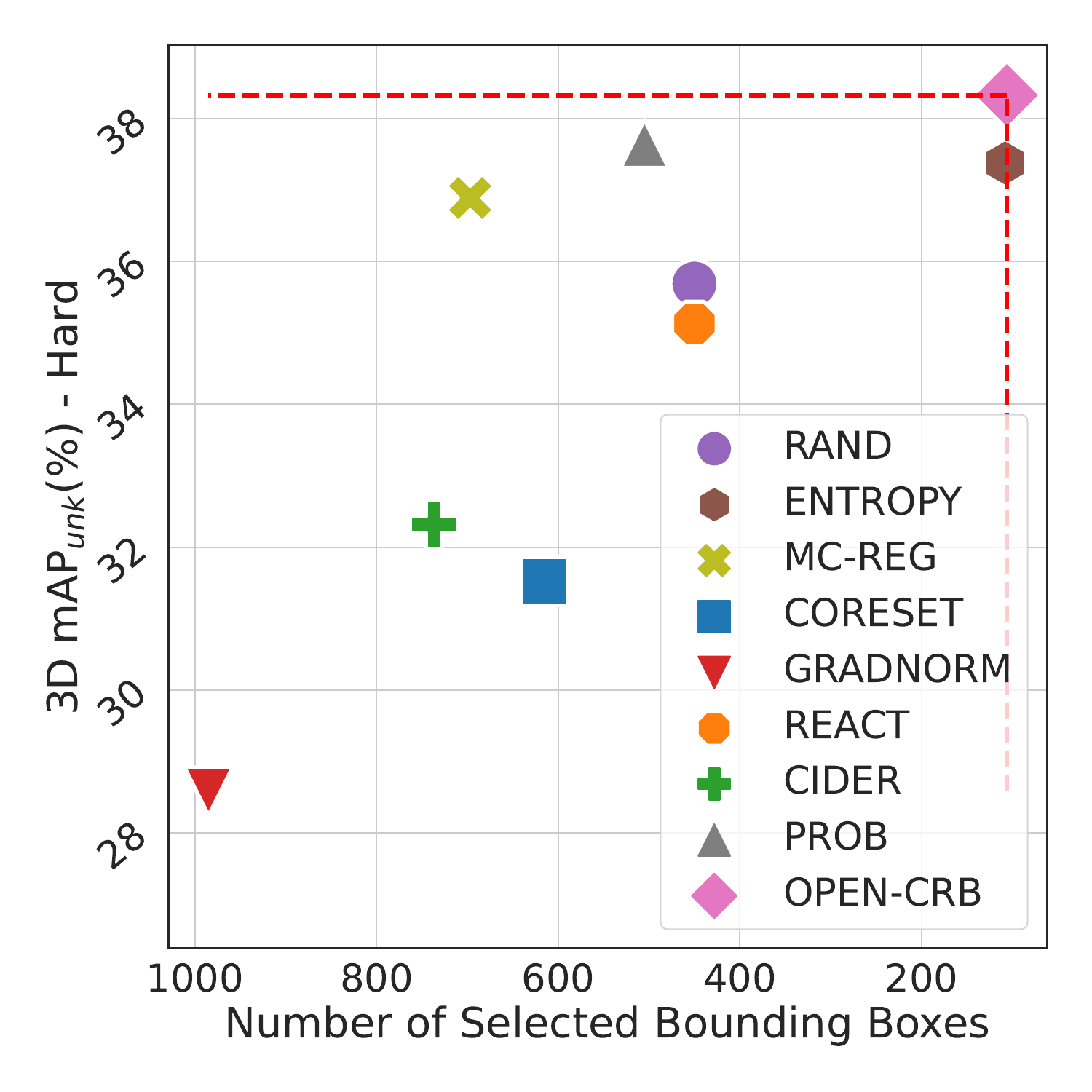} }}%
    \subfloat{{\includegraphics[width=0.245\textwidth]{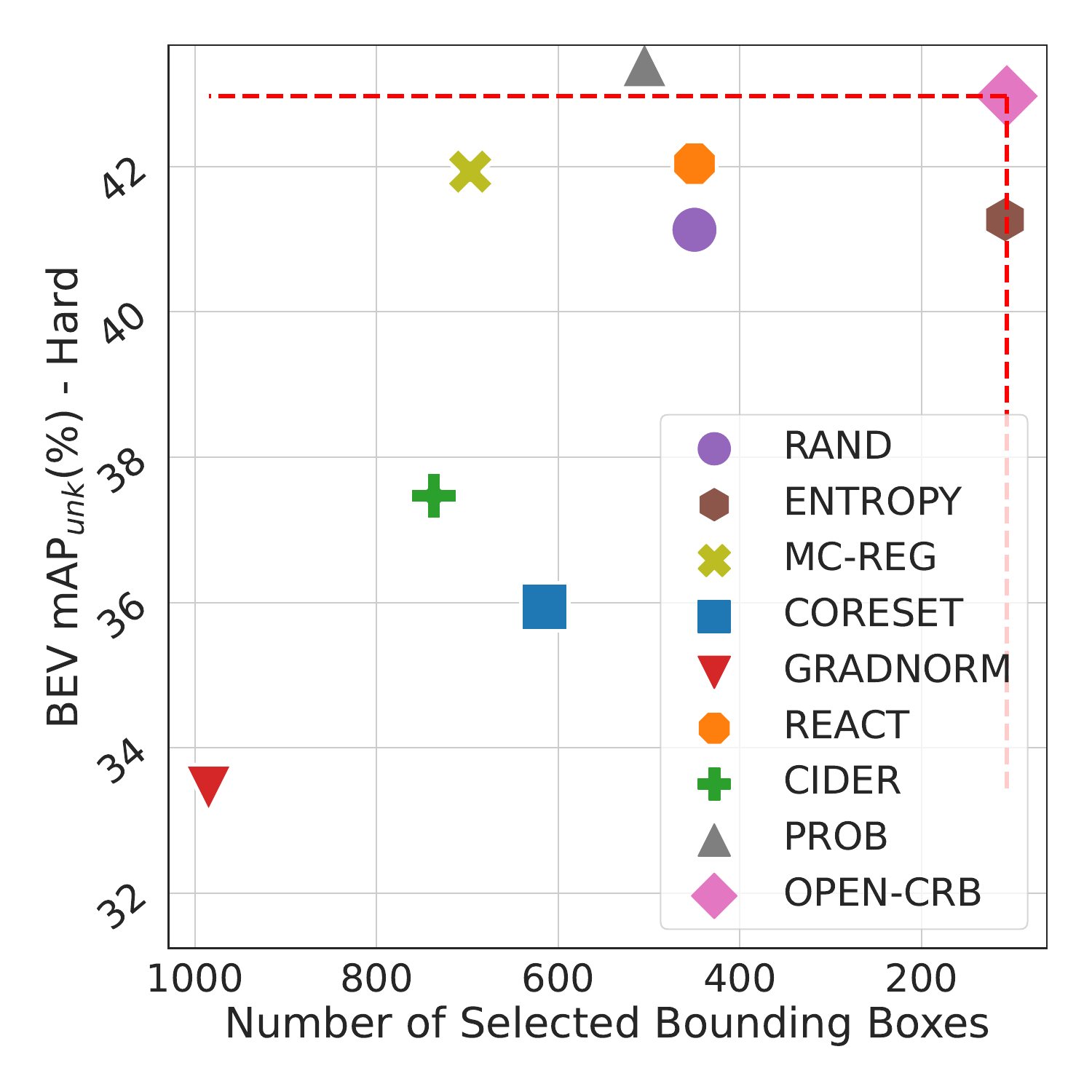} }}%
     \caption{OWAL-3D performance (3D and BEV mAP$_{unk}$ scores) comparisons on \textbf{unknown} classes of {Open-CRB} and baselines on the KITTI dataset. }%
     \label{fig:kitti_owa_unk}
\end{figure}


\begin{figure}[h]%
        \subfloat{{\includegraphics[width=0.245\textwidth]{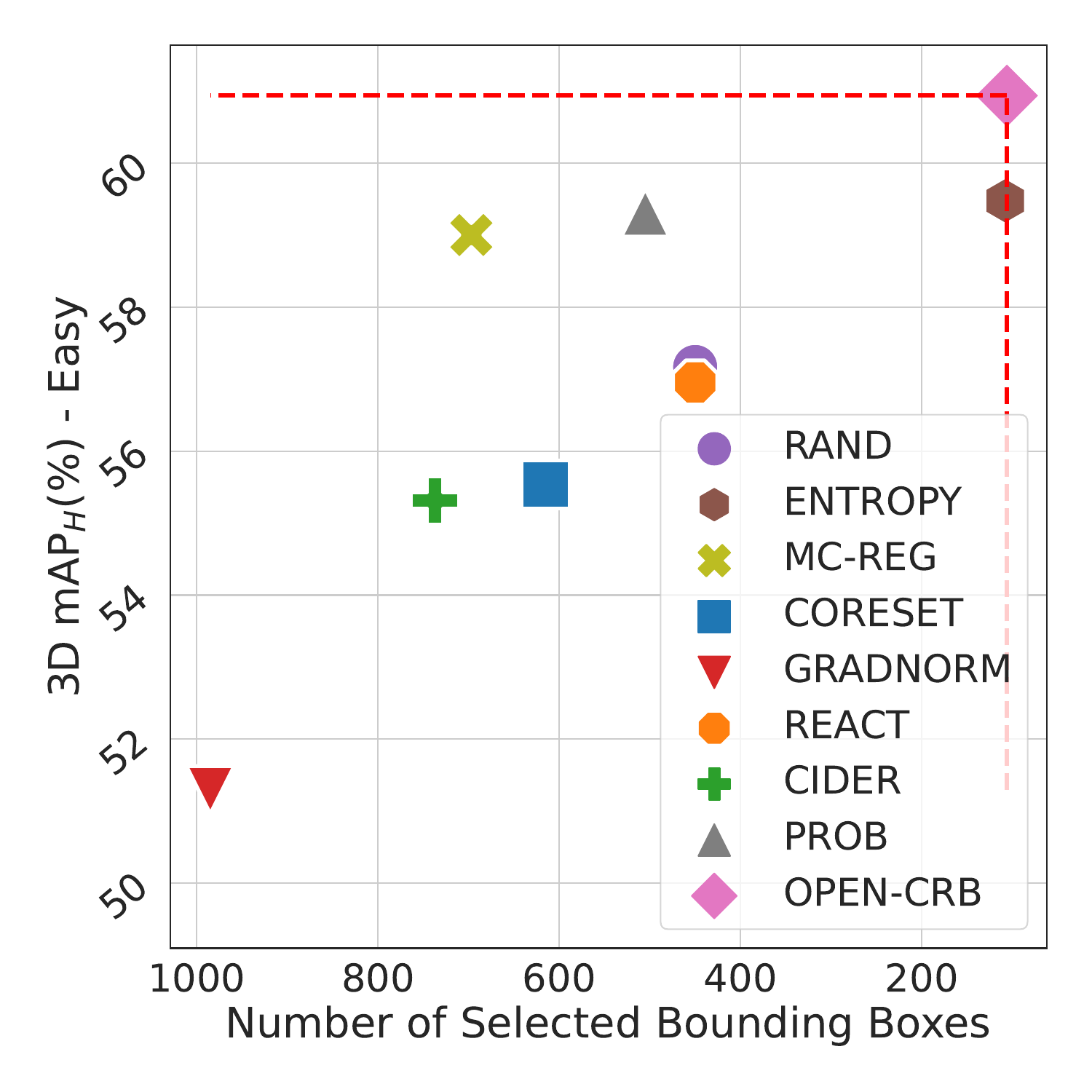} }}%
    \subfloat{{\includegraphics[width=0.245\textwidth]{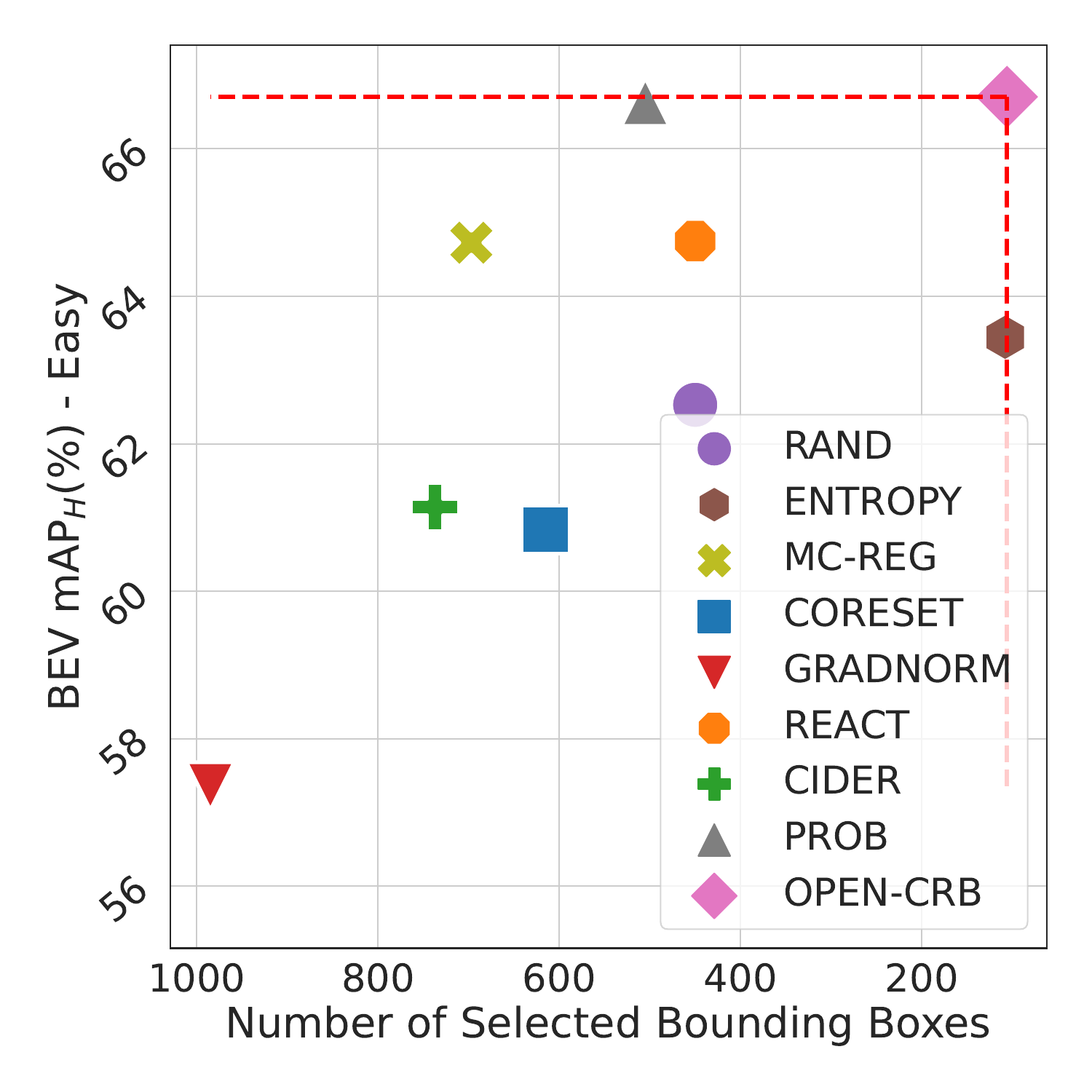} }}%
   \hfill
    \subfloat{{\includegraphics[width=0.245\textwidth]{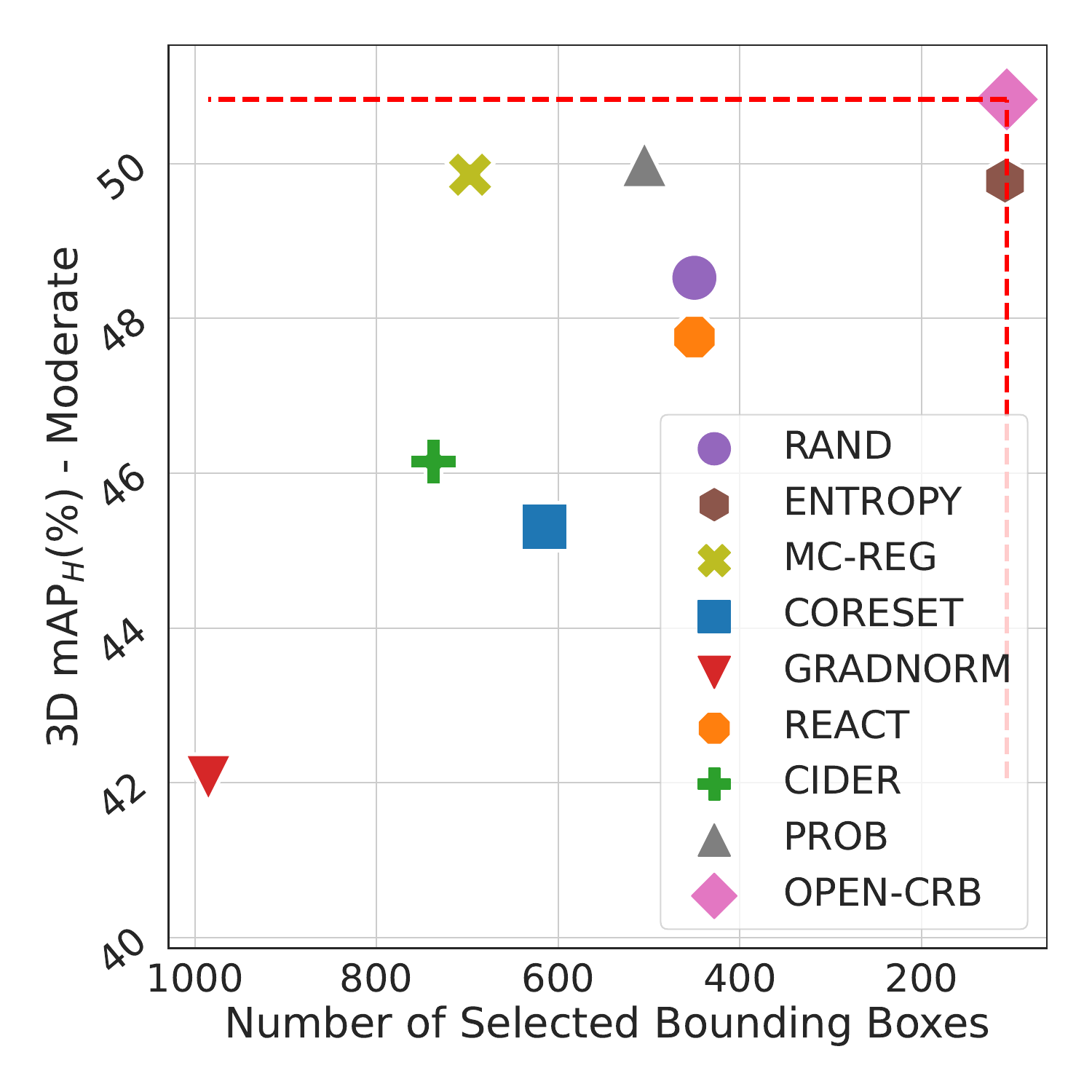} }}%
    \subfloat{{\includegraphics[width=0.245\textwidth]{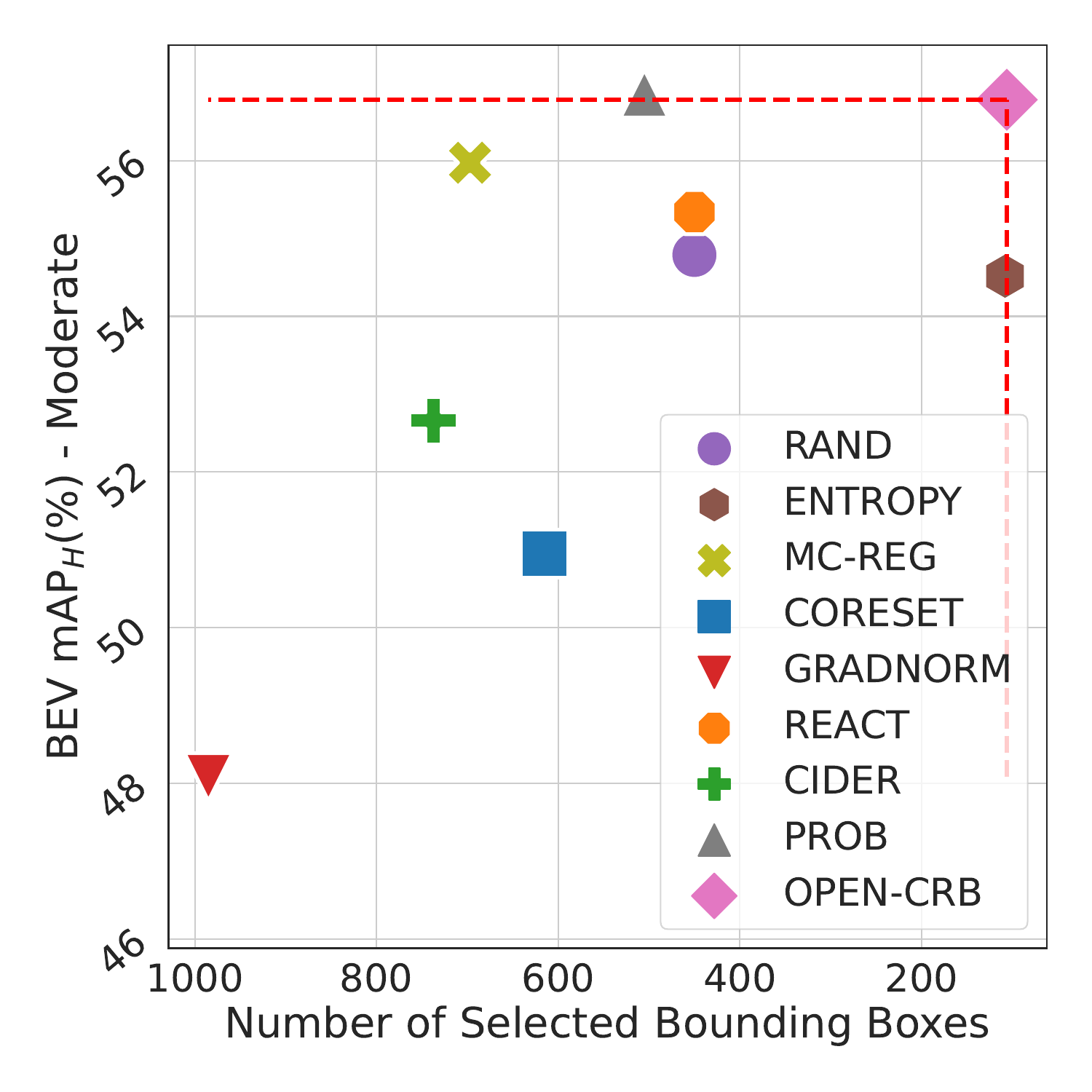} }}%
   \hfill
    \subfloat{{\includegraphics[width=0.245\textwidth]{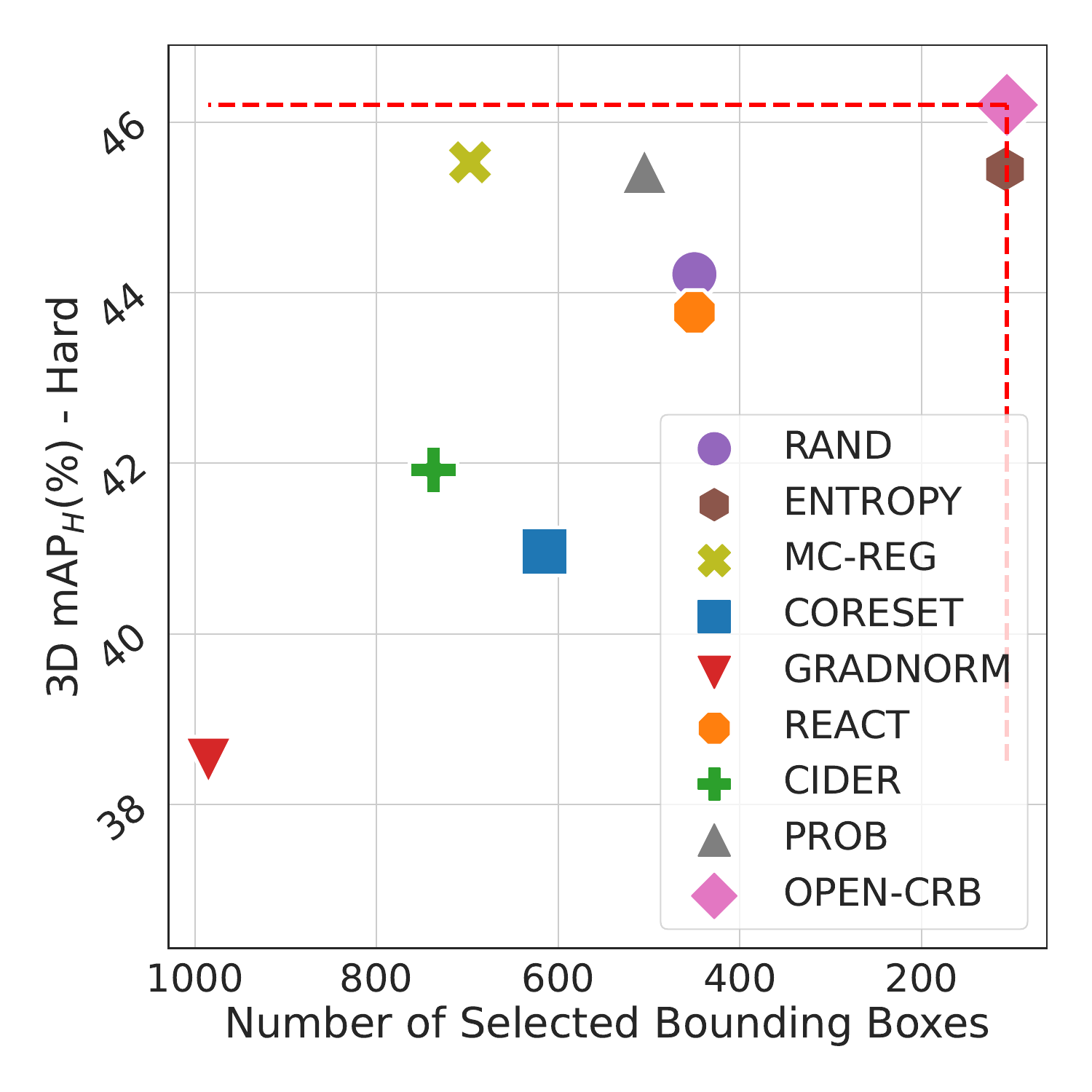} }}%
    \subfloat{{\includegraphics[width=0.245\textwidth]{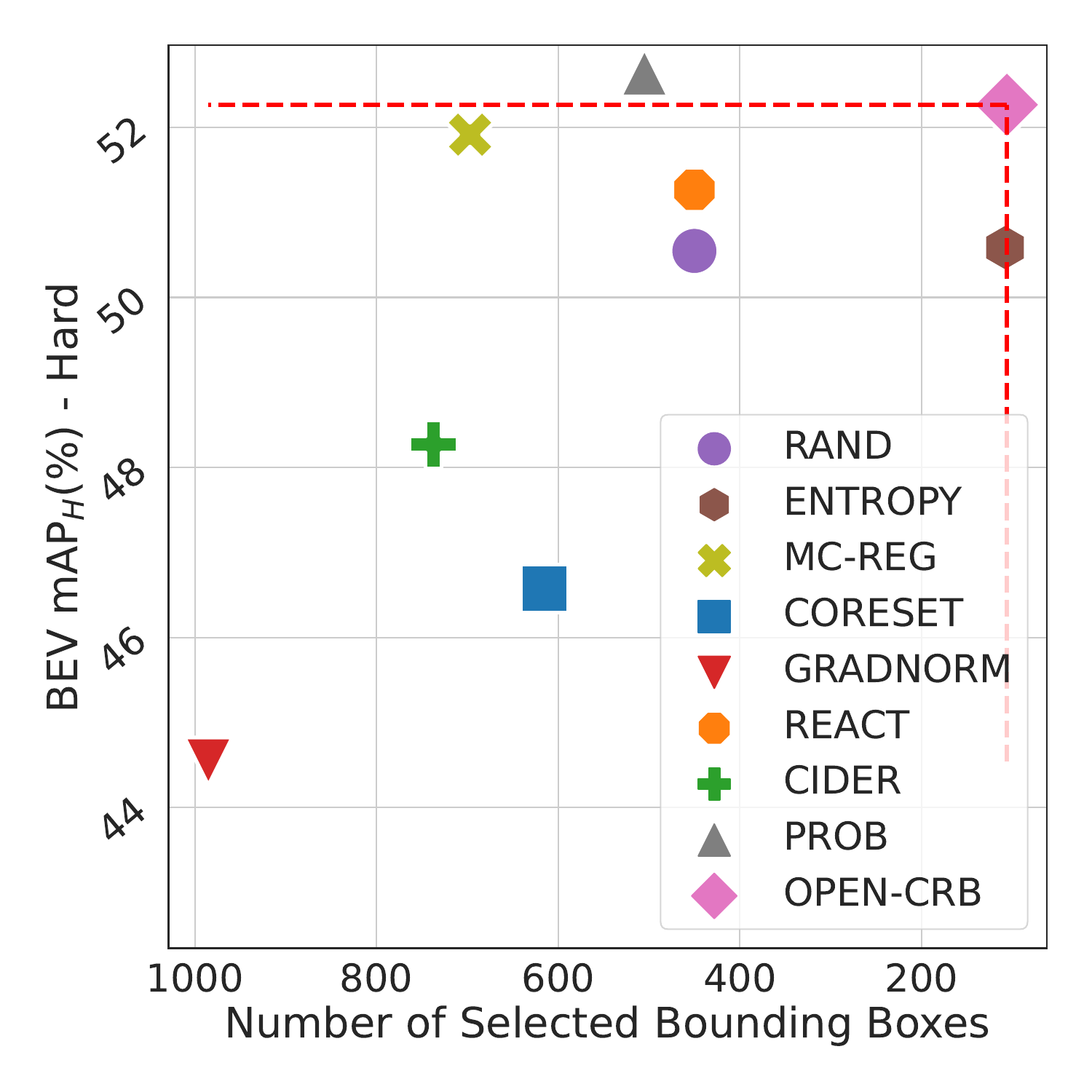} }}%
     \caption{OWAL-3D performance (mAP$_{H}$: 3D and BEV harmonic mean of known mAP and unknown mAP) comparisons on \textbf{all} the classes of {Open-CRB} and baselines on the KITTI dataset.  }%
     \label{fig:kitti_owa_hmean}
\end{figure}

\noindent \textbf{Annotation Costs}. According to the preliminary study \citep{DBLP:conf/iclr/LuoCWYHB23}, we utilize the total number of labeled bounding boxes across all selected point clouds as the unit for realist annotation cost.  

\subsection{Implementation Details}
To ensure the reproducibility of the baselines and the proposed approach, the source code has been made publicly available, including comprehensive training and test configurations, and is readily executable for accessibility and ease of use. For a fair comparison, all methods are constructed from the {PV-RCNN}~\citep{DBLP:conf/cvpr/ShiGJ0SWL20} and {SECOND} \citep{yan2018second} backbones. All experiments are conducted on a GPU cluster with three V100 GPUs. Training {PV-RCNN} on the full set typically requires 20 GPU hours for KITTI and 120 GPU hours for Waymo. While training {SECOND} requires 10 GPU hours for KITTI and 80 GPU hours for nuScenes.

\noindent\textbf{Parameter Settings}. The batch sizes for training and evaluation are fixed to 8, 8, and 16 on KITTI, nuScenes, and Waymo, respectively. The Adam optimizer is adopted with a learning rate initiated as 0.01, and scheduled by one cycle scheduler. The number of {Mc-dropout} stochastic passes is set to 5 for all methods. The $\mathcal{K}_{1}$ and $\mathcal{K}_{2}$ are empirically set to $300$, $200$ for KITTI, $2, 000$ and $1,500$ for nuScenes and $2,000$ and $1,200$ for Waymo. The gradient maps used for ${Rps}$ are extracted from the second convolutional layer in the shared block of the 3D detector. Three dropout layers are enabled during the {Mc-dropout} and the dropout rate is fixed to 0.3. The number of {Mc-dropout} stochastic passes is set to 5 for all methods.

\noindent\textbf{CWAL-3D Protocols}. \textcolor{black}{As our work is the first comprehensive study on active learning for the 3D detection task, the active training protocol for all AL baselines and the proposed method is empirically defined.} For all experiments, we first randomly select $m$ fully labeled point clouds from the training set as the initial $\mathcal{D}_L$. With the annotated data, the 3D detector is trained with $E$ epochs, which is then frozen to select $N_r$ candidates from $\mathcal{D}_U$ for label acquisition. \textcolor{black}{We set the $m$ and $N_r$ to around 3\%  point clouds (\textit{i.e.}, $N_r=m=100$ for KITTI, $N_r=m=400$ for Waymo) to trade-off between reliable model training and high computational costs.} The aforementioned training and selection steps will alternate for $R$ rounds. Empirically, we set $E=30$, $R=6$ for KITTI, and fix $E=40$, $R=5$ for Waymo.

\noindent\textbf{OWAL-3D Protocols}. 
Based on the CWAL-3D protocols, we empirically adopt a similar setup for the OWAL-3D. Specifically, We define $N_r=200$, $m=100$ for KITTI and $N_r=3000$, $m=1500$ for nuScenes. The training and selection steps will alternate for $R$ rounds. Empirically, we set $E=40$, $R=4$ for both KITTI and nuScenes. Regarding the known and unknown class separation, we randomly select a subset of common object categories (\textit{i.e.} car, motorcycle, bicycle, pedestrian, and truck) as known classes in nuScenes dataset. The rest of classes (\textit{i.e.} construction vehicle, bus, trailer, barrier, and traffic cone) are unknown. For the KITTI dataset with only three classes, we set the car and bicycle as known classes and the pedestrian as unknown.

\subsection{Baselines}

We introduce a large-scale open-source benchmark \footnote{accessible at https://github.com/Luoyadan/CRB-active-3Ddet} for both CWAL-3D and OWAL-3D, featuring thorough evaluations, comprehensive analyzes, and 14 extensive cutting-edge baseline algorithms of AL, OW-OD and out-of-distribution (OOD):

\noindent \textbf{Generic Active Learning Baselines:}

\noindent (1) \textbf{{Rand}}: is a basic sampling method that selects $N_r$ samples at random for each selection round; 

\noindent (2) \textbf{{Entropy}} \citep{DBLP:conf/ijcnn/WangS14}: is an \textit{uncertainty}-based active learning approach that targets the \textit{classification} head of the detector, and selects the top $N_r$ ranked samples based on the entropy of the sample's predicted label; 

\noindent (3) \textbf{{LLAL}} \citep{DBLP:conf/cvpr/YooK19}: is an \textit{uncertainty}-based method that adopts an auxiliary network to predict an indicative loss and enables to select samples for which the model is likely to produce wrong predictions; 

\noindent (4) \textbf{{Coreset}} \citep{DBLP:conf/iclr/SenerS18}: is a \textit{diversity}-based method performing the core-set selection that uses the greedy furthest-first search on both labeled and unlabeled embeddings at each round;  

\noindent (5) \textbf{{Badge}} \citep{DBLP:conf/iclr/AshZK0A20}: is a \textit{hybrid} approach that samples instances that are disparate and of high magnitude when presented in a hallucinated gradient space.

\noindent (6) \textbf{{Bait}} \citep{DBLP:conf/nips/AshGKK21}: selects batches of samples by optimizing a bound on the maximum likelihood estimators (MLE) error in terms of the Fisher information.

\noindent  \textbf{Applied AL Baselines for 2D and 3D Detection}: for a fair comparison, we also compared three variants of the deep active learning method for 3D detection and adapted one 2D active detection method to our 3D detector. 

\noindent (7) \textbf{{Mc-mi}} \citep{DBLP:conf/ivs/FengWRMD19} utilized Monte Carlo dropout associated with mutual information to determine the uncertainty of point clouds.


\noindent (8) \textbf{{Mc-reg}}: additionally, to verify the importance of the uncertainty in regression, we design an \textit{uncertainty}-based baseline that determines the \textit{regression} uncertainty via conducting $M$-round {Mc-dropout} stochastic passes at the test time. The variances of predictive results are then calculated, and the samples with the top-$N_r$ greatest variance will be selected for label acquisition. We further adapted two applied AL methods for 2D detection to a 3D detection setting, where 

\noindent (9) \textbf{{Lt/c}} \citep{DBLP:conf/accv/KaoLS018} measures the class-specific localization tightness, \textit{i.e.}, the changes from the intermediate proposal to the final bounding box and 

\noindent (10) \textbf{{Consensus}} \citep{DBLP:conf/ivs/SchmidtRTK20} calculates the variation ratio of minimum IoU value for each RoI-match of 3D boxes.

\noindent \textbf{Applied AL Baselines for OW-OD and OOD:} although existing OW-OD methods follows a different paradigms (\textit{i.e.}, out-of-distribution detection (OOD) / open-set recognition plus continual learning), we can still implement unknown exploration modules to detect potential unknown objects in 3D scenes. Thus, the AL strategy becomes selecting point clouds with higher probability to contain unknown objects. 

\noindent(11) \textbf{{PROB}} \citep{DBLP:conf/cvpr/ZoharWY23} integrates probabilistic models into the object detector to facilitate objectiveness estimation within the embedded feature space. Then, point clouds containing higher averaged estimated objectiveness are selected. 

\noindent(12) \textbf{{GradNorm}} \citep{DBLP:conf/nips/HuangGL21} splits in-distribution (ID) and OOD data based on the vector norm of gradients, backpropagated from the discrepancy between the softmax output and a uniform probability distribution. 

\noindent (13) \textbf{{ReAct}} \citep{liu2020energy_ood} separates the ID and OOD data after rectifying the activations at an upper limit, based on the observation that OOD data have larger variations in activations. 

\noindent (14) \textbf{{Cider}} \citep{DBLP:conf/iclr/MingSD023} formalizes the latent representations as vMF distributions, then calculate distances in hyperspherical embeddings, from data point to the class prototypes.

\subsection{Main Results for OWAL-3D}\label{sec:main_results_OWAL-3D}


\begin{figure*}[!ht]%
    \subfloat{{\includegraphics[width=0.29\textwidth]{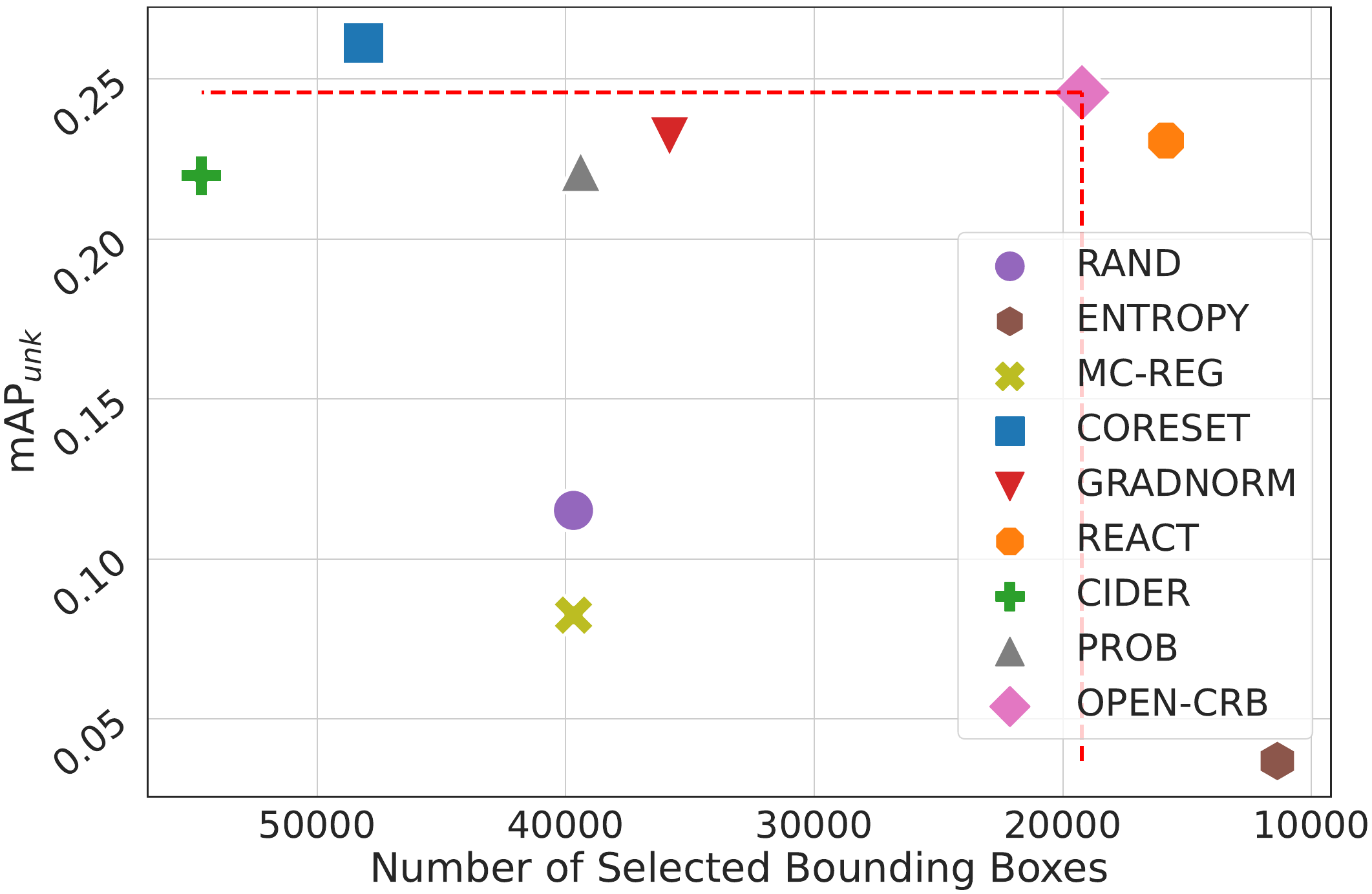} }}%
    \subfloat{{\includegraphics[width=0.29\textwidth]{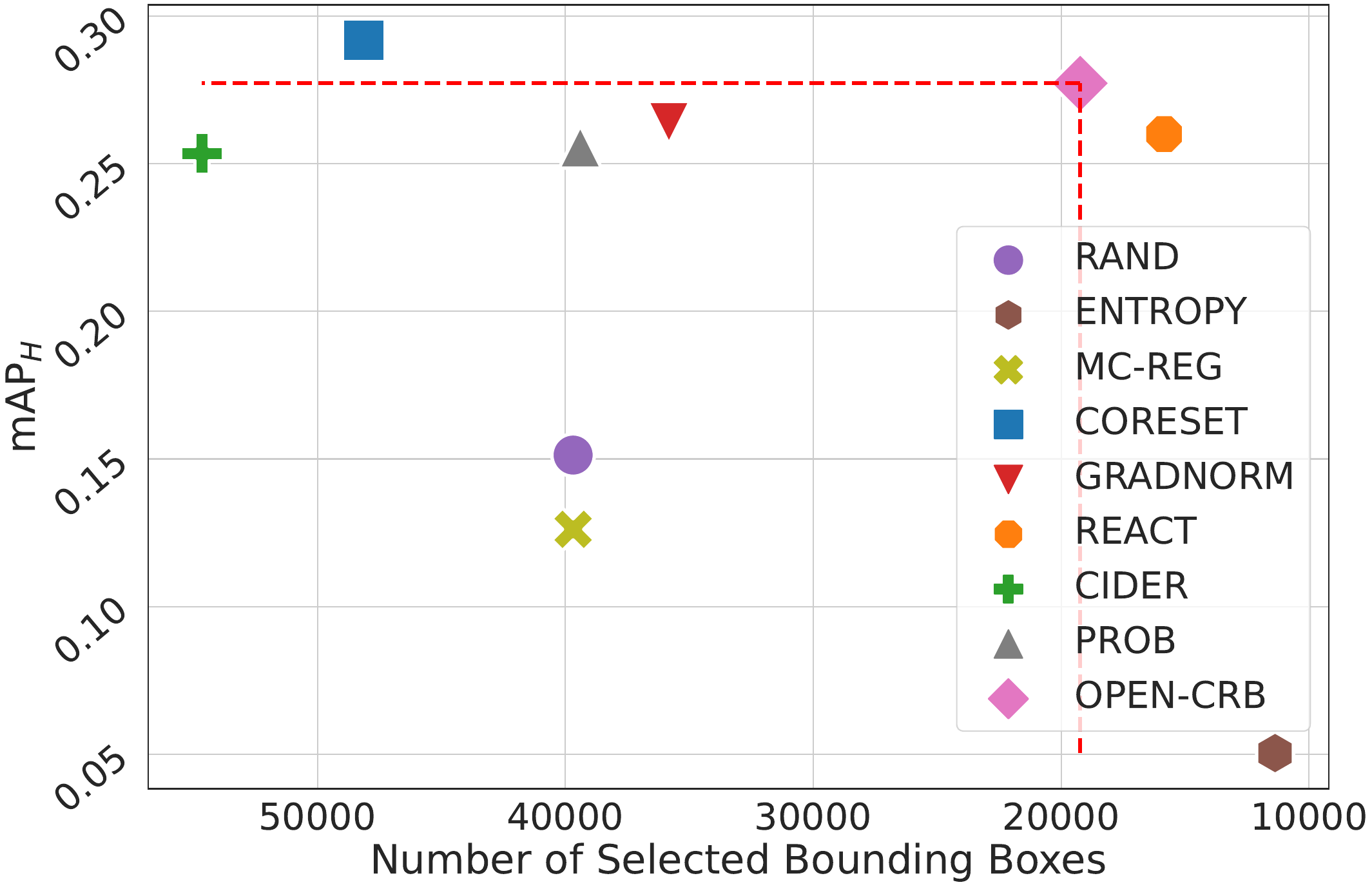} }}%
\subfloat{{\includegraphics[width=0.41\textwidth]{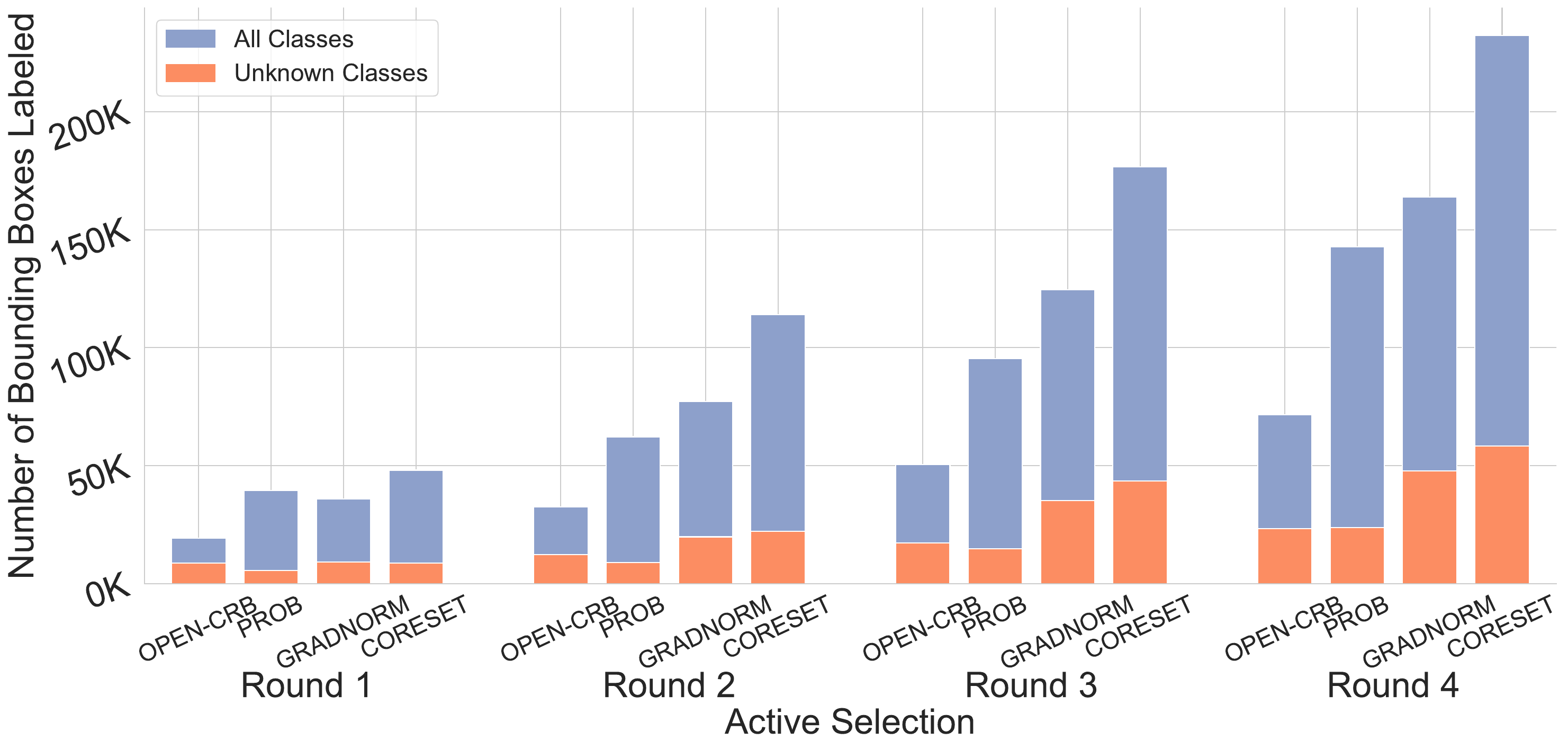} }}\vspace{-2ex}
\caption{\textbf{Left} (two scatter plots): OWAL-3D performance (mAP$_{unk}$ and mAP$_{H}$) comparisons of {Open-CRB} and baselines on the nuScenes dataset. \textbf{Right} (bar plot): The accumulation of the number of selected bounding boxes from nuScenes dataset, with active learning selection rounds increase, under the OWAL-3D setting. }%
\label{fig:nuscenes_perf_and_select_count}
\end{figure*}

\begin{figure*}[!ht]%
\subfloat{{\includegraphics[width=0.19\textwidth]{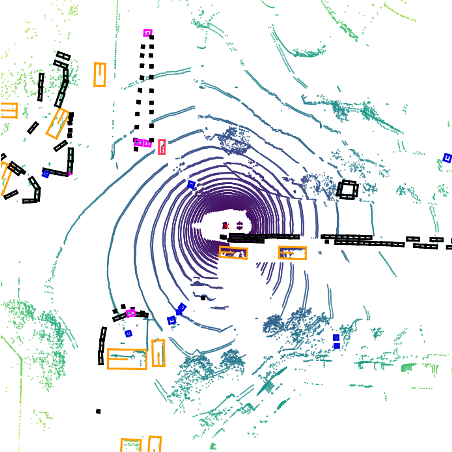} }}%
\subfloat{{\includegraphics[width=0.19\textwidth]{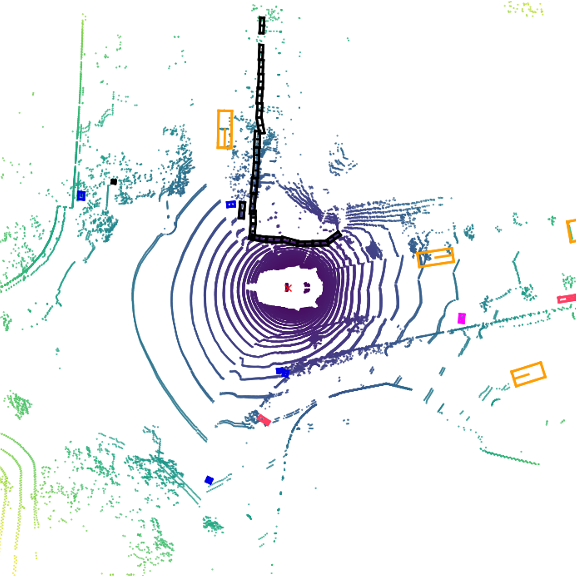} }}%
\subfloat{{\includegraphics[width=0.19\textwidth]{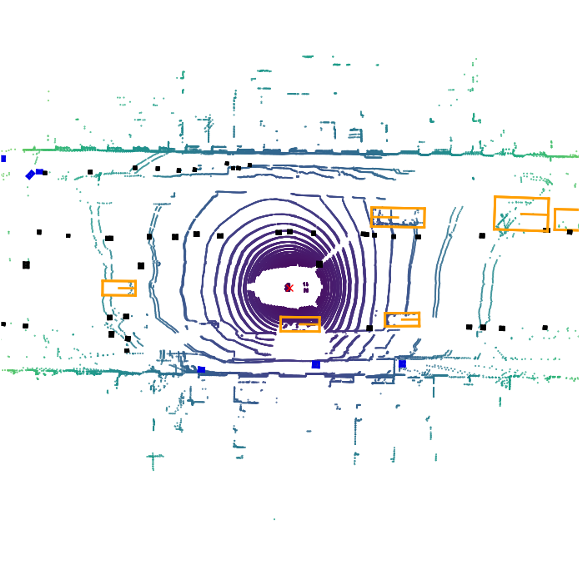} }}%
\subfloat{{\includegraphics[width=0.19\textwidth]{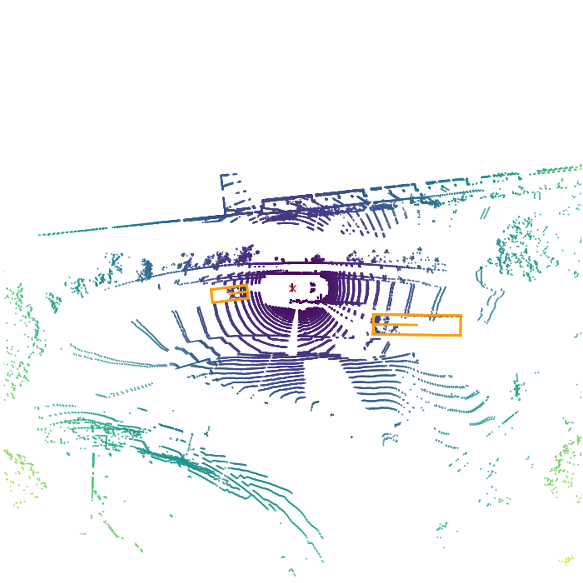} }}%
\subfloat{{\includegraphics[width=0.19\textwidth]{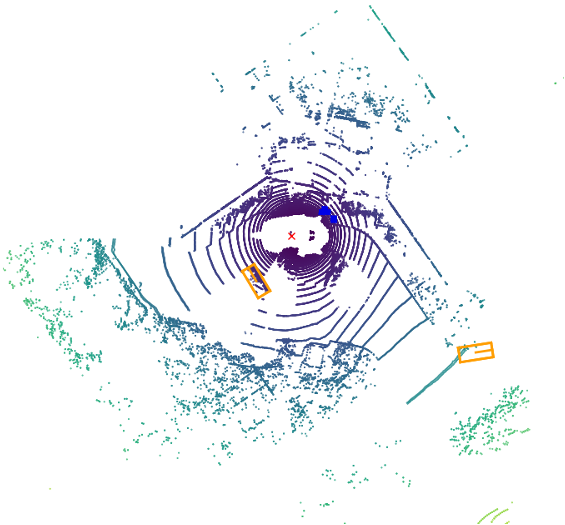} }}\vspace{-2ex}
\caption{Visualization of acquired point clouds by the proposed Open-CRB. The inverse relationship of OLC led to the selection of point clouds with either a large number of instances from novel categories (first three) or very few known classes (last two) to minimize annotation costs. The harmonic relationship ensures that object categories are diverse across all five point clouds.}%
\label{fig:kitti_owa_vis_CRB}
\end{figure*}

\begin{table}[t] \label{fig:compare_to_crb}
\centering 
\caption{OWAL-3D performance (3D mAP scores) comparisons when incorporating the proposed OLC to generic AL methods on the KITTI \textit{val} set with 1000 queried bounding boxes. The best results are highlighted in bold, and the second-best results are underlined.}
\resizebox{1\linewidth}{!}{%
\begin{tabular}{l l l l l l}
\toprule 
& Methods &\multicolumn{1}{c}{{3D {mAP}$_{unk}$}}&\multicolumn{1}{c}{{BEV {mAP}$_{unk}$}}&\multicolumn{1}{c}{{3D {mAP}}} & \multicolumn{1}{c}{{BEV {mAP}}} \\ 
\midrule
&{Random} & 43.05 & 48.62 & 56.09 & 63.22 \\
&{Random + OLC} & 44.44$_{\textcolor{red}{\text{3.23\%}\uparrow}}$ & 50.68$_{\textcolor{red}{\text{4.24\%}\uparrow}}$ & 57.95$_{\textcolor{red}{\text{3.32\%}\uparrow}}$ & 65.20$_{\textcolor{red}{\text{3.13\%}\uparrow}}$ \\
\midrule
&{Cider} & 36.23 & 41.91 & 54.71 & 62.04 \\
&{Cider + OLC} & 39.89$_{\textcolor{red}{\text{10.10\%}\uparrow}}$ & 46.16$_{\textcolor{red}{\text{10.14\%}\uparrow}}$ & 54.63$_{\textcolor{red}{\text{0.15\%}\downarrow}}$ & 62.22$_{\textcolor{red}{\text{0.29\%}\uparrow}}$ \\
\midrule
&{GradNorm} & 31.43 & 36.34 & 52.85 & 59.47 \\
&{GradNorm + OLC} & 42.66$_{\textcolor{red}{\text{35.73\%}\uparrow}}$ & 47.32$_{\textcolor{red}{\text{30.21\%}\uparrow}}$ & 57.80$_{\textcolor{red}{\text{9.37\%}\uparrow}}$ & 63.74$_{\textcolor{red}{\text{7.18\%}\uparrow}}$ \\
\midrule
&{CRB} & \underline{46.42} & \underline{51.36} & \underline{59.22} & \underline{66.51} \\
&{Open-CRB} & \textbf{49.23}$_{\textcolor{red}{\text{6.05\%}\uparrow}}$ & \textbf{55.17}$_{\textcolor{red}{\text{7.42\%}\uparrow}}$ & \textbf{60.83}$_{\textcolor{red}{\text{2.72\%}\uparrow}}$ & \textbf{67.71}$_{\textcolor{red}{\text{1.80\%}\uparrow}}$ \\
\bottomrule 
\end{tabular}
}
\label{tab:abl_crb_vs_open_crb} 
\end{table}

\begin{figure*}[!ht]
\centering
\includegraphics[width=1\linewidth]{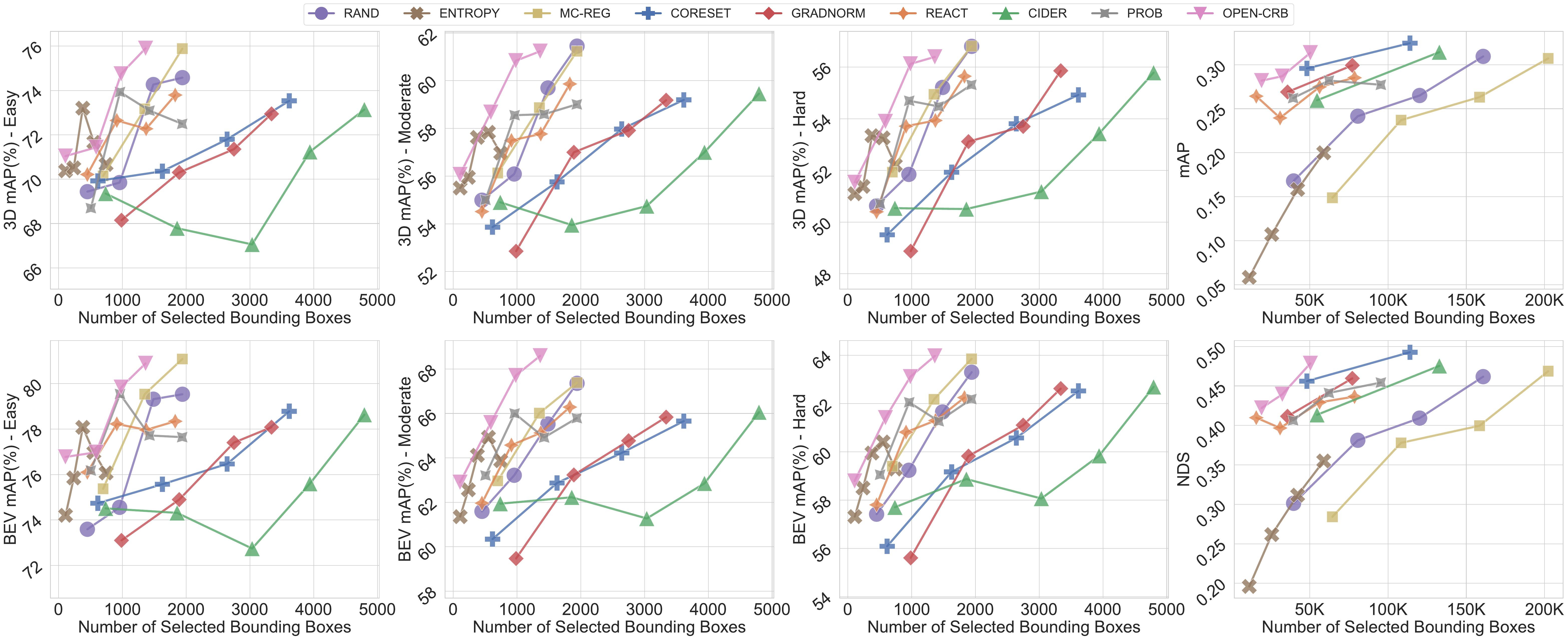}
\caption{OWAL-3D performance (3D and BEV mAP scores) comparisons of {Open-CRB} and AL baselines on the KITTI (first three columns) and nuScenes (final column) datasets, with increasing annotation cost. }
\label{fig:OWAL-3D_AL_KITTI}
\end{figure*}

We performed extensive experiments on both the KITTI and nuScenes datasets to validate the efficacy of the proposed {Open-CRB}, utilizing {SECOND} as the backbone detector. We illustrate the relationship between annotation cost and the corresponding performance improvement through scatter plots, as depicted in Figure \ref{fig:kitti_owa_unk}, Figure \ref{fig:kitti_owa_hmean} and Figure \ref{fig:nuscenes_perf_and_select_count}. 

\noindent  \textbf{Results on Unknown Classes.} It is worth noting that the proposed {Open-CRB} suggested significantly surpasses other baseline techniques in enhancing the capability to recognize unknown classes after the first selection is found. For instance, the upper three plots in Figure \ref{fig:kitti_owa_unk} feature a horizontal dashed line, representing the best 3D mAP$_{unk}$ achieved by {Open-CRB} for recognizing unknown classes. The most inspiring finding is that the group of uncertainty-based baselines (\textit{e.g.}, Entropy, PROB and MC-Reg) achieve better mAP$_{unk}$ than discrepancy-based methods (\textit{e.g.}, Coreset and Cider). This finding validates the effectiveness of uncertainty for learning unknown class and thus become the basis of our method. Turning to the nuScenes dataset, the left plot in Figure \ref{fig:kitti_owa_hmean} demonstrates that {Open-CRB} secures the second-highest mAP$_{unk}$. Although {Coreset} achieves a slightly superior result, \textbf{it comes at the expense of 2.5 times the annotation cost when compared to {Open-CRB}}. Moreover, in the context of nuScenes, discrepancy-based methods (\textit{e.g.}, Coreset and Cider) produce similar results to those of uncertainty-based baselines, however, at the expense of selecting a significantly higher number of bounding boxes. Overall, the impressive results demonstrated by the {Open-CRB} in unknown categories provide strong evidence that the proposed sampling technique is highly effective in selecting informative objects from novel categories within unlabeled point clouds. This approach significantly enhances the model's ability to adapt to open world scenarios. 

\noindent \textbf{Results on All Classes.} To evaluate the effectiveness of the proposed methods in recognizing both known and unknown classes, we adopt the widely used harmonic mean Average Precision (mAP$_H$) balanced between mAP$_{unk}$ and mAP$_{k}$, and present the results in Figure \ref{fig:kitti_owa_hmean} for KITTI and Figure \ref{fig:nuscenes_perf_and_select_count} for nuScenes. A higher mAP$_H$ indicates that the method not only performs better across all categories but also maintains a narrow gap between mAP$_{unk}$ and mAP$_{k}$. In the case of KITTI, as depicted in Figure \ref{fig:kitti_owa_hmean}, the proposed {Open-CRB} consistently achieves the highest 3D mAP$_H$ across all difficulty settings compared to other baselines, requiring the least annotation cost. The very recent state-of-the-art OW-OD method, PROB, achieves a slightly higher BEV mAP$_H$ than {Open-CRB} (52.64\% vs. 52.27\%), but it costs significantly more annotated bounding boxes (505 vs. 106). Besides, PROB is not a post-hoc approach, as it requires additional computations at the pre-training phase, leading to more time consumption than {Open-CRB}. In the context of the more challenging label-rich dataset, nuScenes, {Open-CRB} continues to perform remarkably while incurring very limited labeling cost (19232 of labeled bounding boxes), securing a mAP$_H$ that is second only to {Coreset} (48133 of labeled bounding boxes). These experimental results clearly demonstrate that our approach effectively enables the model to simultaneously learn both unknown and known classes without bias.

\noindent \textbf{Annotation Cost and Performance Balance.} To assess the annotation cost across various approaches, we present the total count of labeled bounding boxes (including both known and unknown classes) on the x-axis. It is clear that our approach stands out by significantly reducing the total number of annotations required (106 from KITTI and 8734 from nuScenes) compared to the majority of baseline methods, all while preserving the outstanding $\text{mAP}_{unk}$ and $\text{mAP}_{H}$ score. This holds especially true for KITTI, as evident from the scatter plots in Figure \ref{fig:kitti_owa_hmean}, our approach successfully reaches the \textbf{skyline point}, representing the optimum in both the dimension of performance and the dimension of annotation costs. 

\noindent \textbf{Results with Increasing Annotations.} Additionally, we illustrate the performance trends in Figure \ref{fig:OWAL-3D_AL_KITTI} for KITTI and nuScenes, respectively, depicting how performance evolves as the labeling costs increase during multi-round active selections and model training. We can clearly observe that {Open-CRB} consistently surpasses all state-of-the-art methods by a significant margin, regardless of the quantity of annotated bounding boxes, difficulty settings in KITTI or the evaluation metric used in nuScenes. It is remarkable that, on the nuScenes dataset, the annotation time for the proposed {Open-CRB} is three times quicker than {Rand} (approximately 50,000 annotations versus approximately 160,000 at round 3), while achieving comparable performance. 

\noindent \textbf{Analysis on Selected Labels.}
The experimental analysis above demonstrates the effectiveness of our proposed method in enhancing model performance. This success has sparked our curiosity about the specific source of performance improvement, in other words, the types of point clouds the model trains on to achieve these outstanding results. To investigate this, we delved into the composition of known and unknown labels within selected boxes using different methods, illustrated in the bar plot of Figure \ref{fig:nuscenes_perf_and_select_count}. We tracked the accumulated count of known and unknown boxes in the nuScenes dataset as the active selection round increases. Note that, while unseen classes were labeled after round 1, we continued to track them in subsequent active rounds. It is clear from the initial round, our method selected the fewest total boxes while the highest proportion of unknown instances: 45.41\% of which belonged to unknown classes. In contrast, other methods not only incur higher costs but also fail to mine point clouds containing novel class objects. Specifically, for {PROB}, {GradNorm}, and {Coreset}, the percentages of selected unknown classes objects in the first round were 14.01\%, 25.42\%, and 18.25\%, respectively. This highlights our approach as a targeted solution to the core challenge of OWAL-3D task: how to acquire point clouds housing potential novel class objects while minimizing annotation costs. In subsequent rounds, {Open-CRB} which reverts to CRB, consistently maintains a high percentage of boxes belonging to these newly introduced classes because CRB is able to seek diverse known labels. By the final round, the percentages of boxes for new arrival classes are 32.46\%, 16.65\%, 29.24\%, and 25.11\% for each method, respectively. These findings for the composition (unknown / unknown) of the acquired labels serve as direct evidence that the proposed method achieves its intended objectives, as shown in Figure \ref{fig:OLC}.

\noindent \textbf{Analysis on Open Label Conciseness.} In this section, we plug open Label Conciseness (OLC) policy to different closed world AL strategy to evaluate its effectiveness. Specifically, we follow the Open-CRB framework which adopts OLC in the first selection round to maximize knowledge acquisition from novel classes, then in the subsequent rounds, we revert to closed world AL methods, such as random, GradNorm and Cider. As shown in Table \ref{fig:compare_to_crb}, incorporating OLC leads to a significant improvement across all generic AL methods, particularly in detecting objects of novel classes. Notably, GranNorm achieves a 35.73\% improvement in unknown classes when querying 1000 boxes. These findings demonstrate that OLC can select sufficient novel concepts in the first active round to optimize the model, efficiently transforming the task into a closed world scenario where any traditional AL method can be applied.




\noindent \textbf{Qualitative Analysis.} We perform a qualitative analysis of the acquired point clouds to provide a more intuitive understanding of the advantages of the proposed OLC selection policy. The point clouds sampled by OLC in the first round are illustrated in Figure \ref{fig:kitti_owa_vis_CRB}. In the first three frames, we observe a wide range of previously unseen categories, such as barriers and traffic cones. In contrast, the last two frames contain only a few representative known instances. This outcome benefits from the inverse relationship (Figure \ref{fig:OLC} and Eq. \eqref{eq:inverse}) of OLC, which enhances the potential to select point clouds with novel classes. Moreover, the object categories across all selected frames are concise and diverse, which is consistent with the harmonic relationship (Figure \ref{fig:OLC} and Eq. \eqref{eq:harmonic}), effectively reducing redundancy.

\subsection{Main Results for CWAL-3D}
\begin{figure}[!ht]%
\subfloat{{\includegraphics[width=0.245\textwidth]{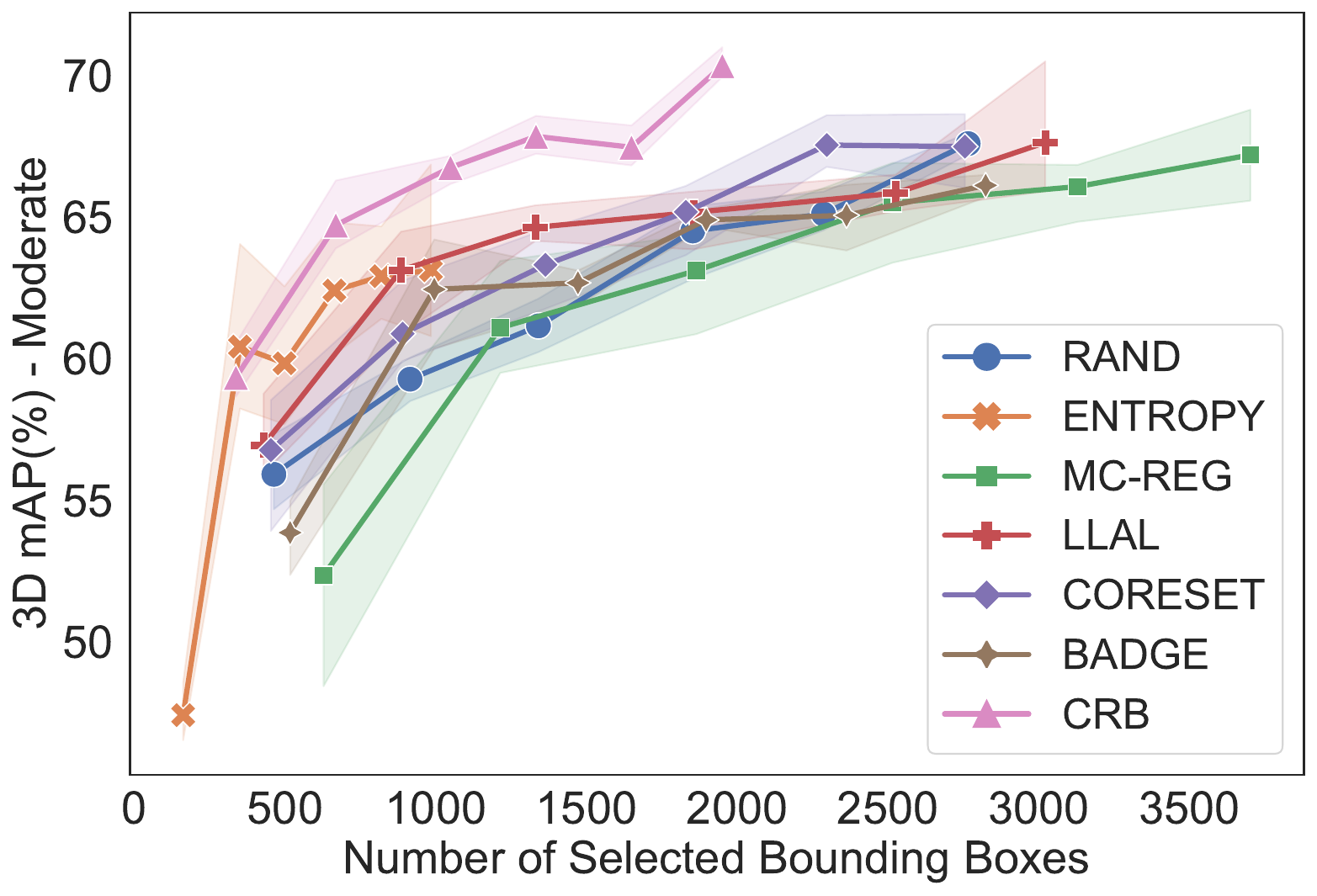} }}%
    \subfloat{{\includegraphics[width=0.245\textwidth]{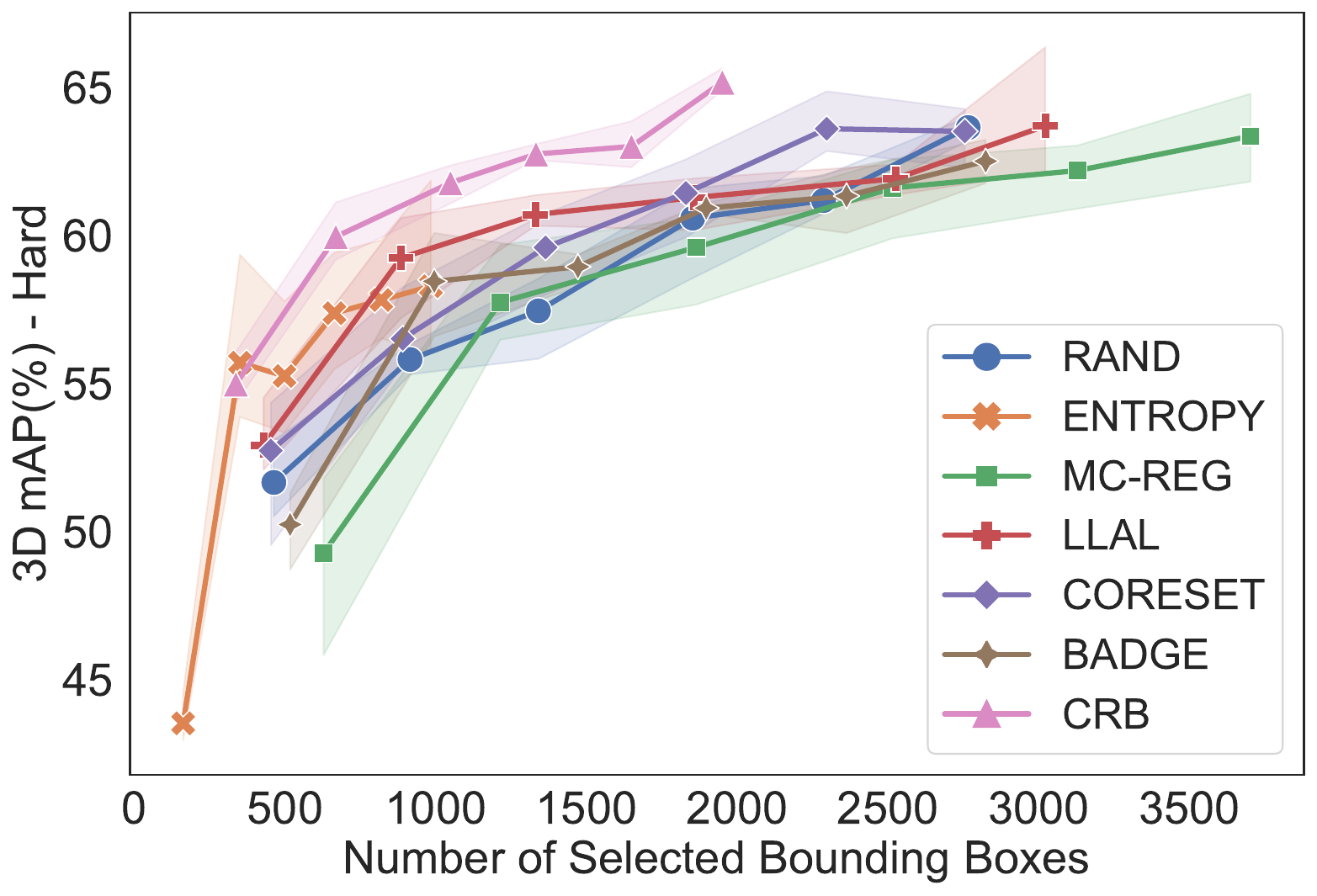} }}%

   \hfill
    \subfloat{{\includegraphics[width=0.245\textwidth]{Figures/kitti_3d_hard_box_level.pdf} }}%
    \subfloat{{\includegraphics[width=0.245\textwidth]{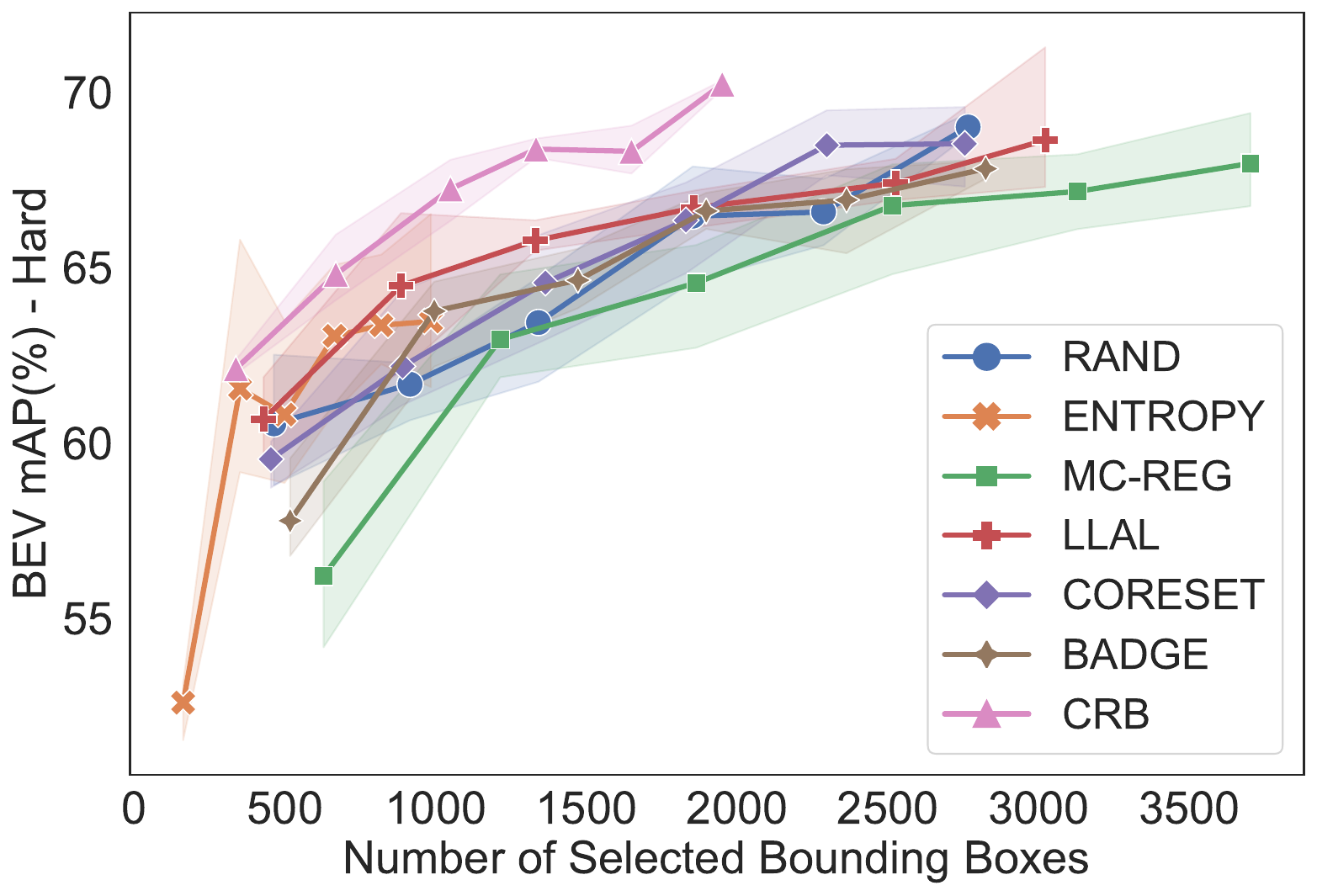} }}%
   \hfill
    \subfloat{{\includegraphics[width=0.245\textwidth]{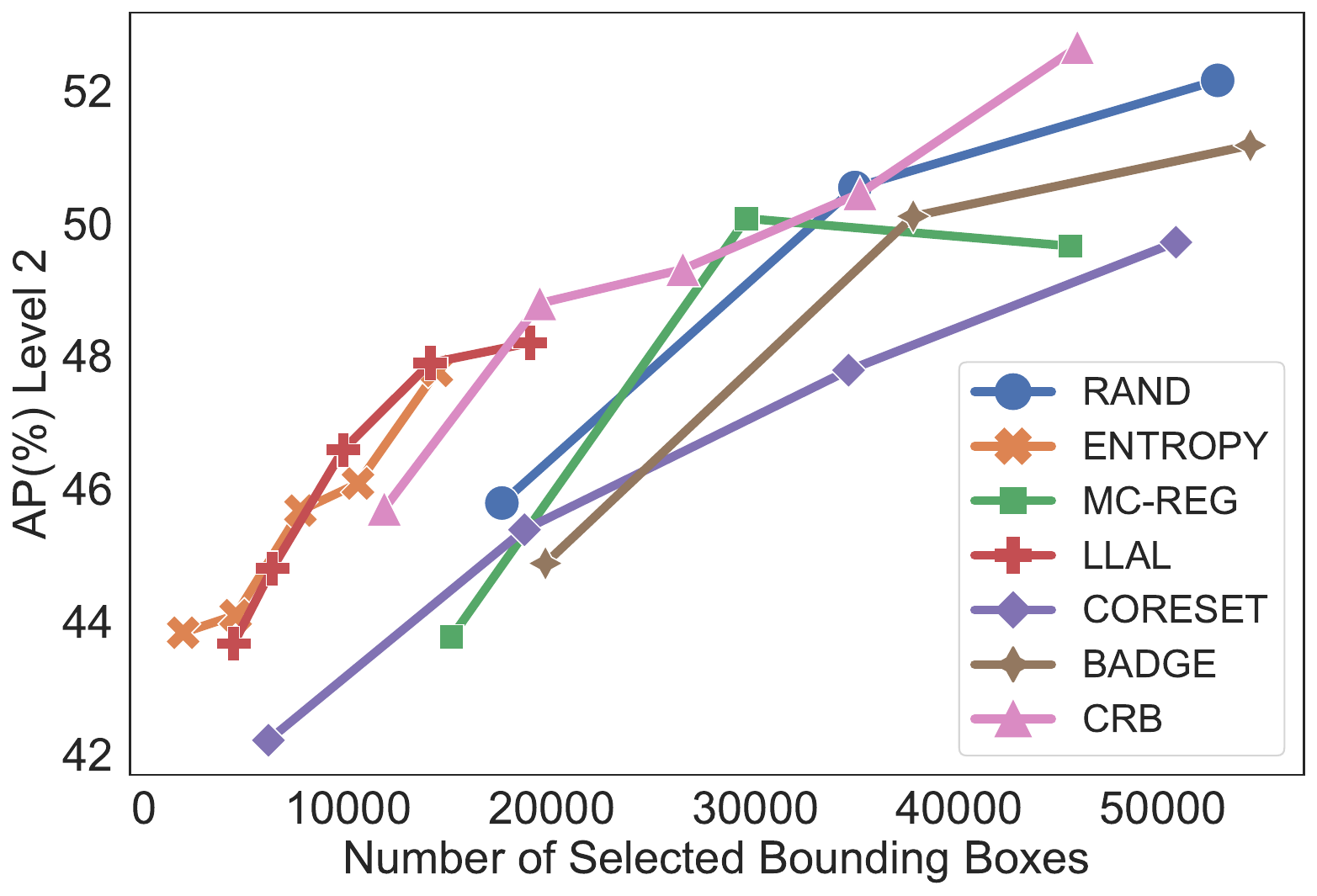} }}%
    \subfloat{{\includegraphics[width=0.245\textwidth]{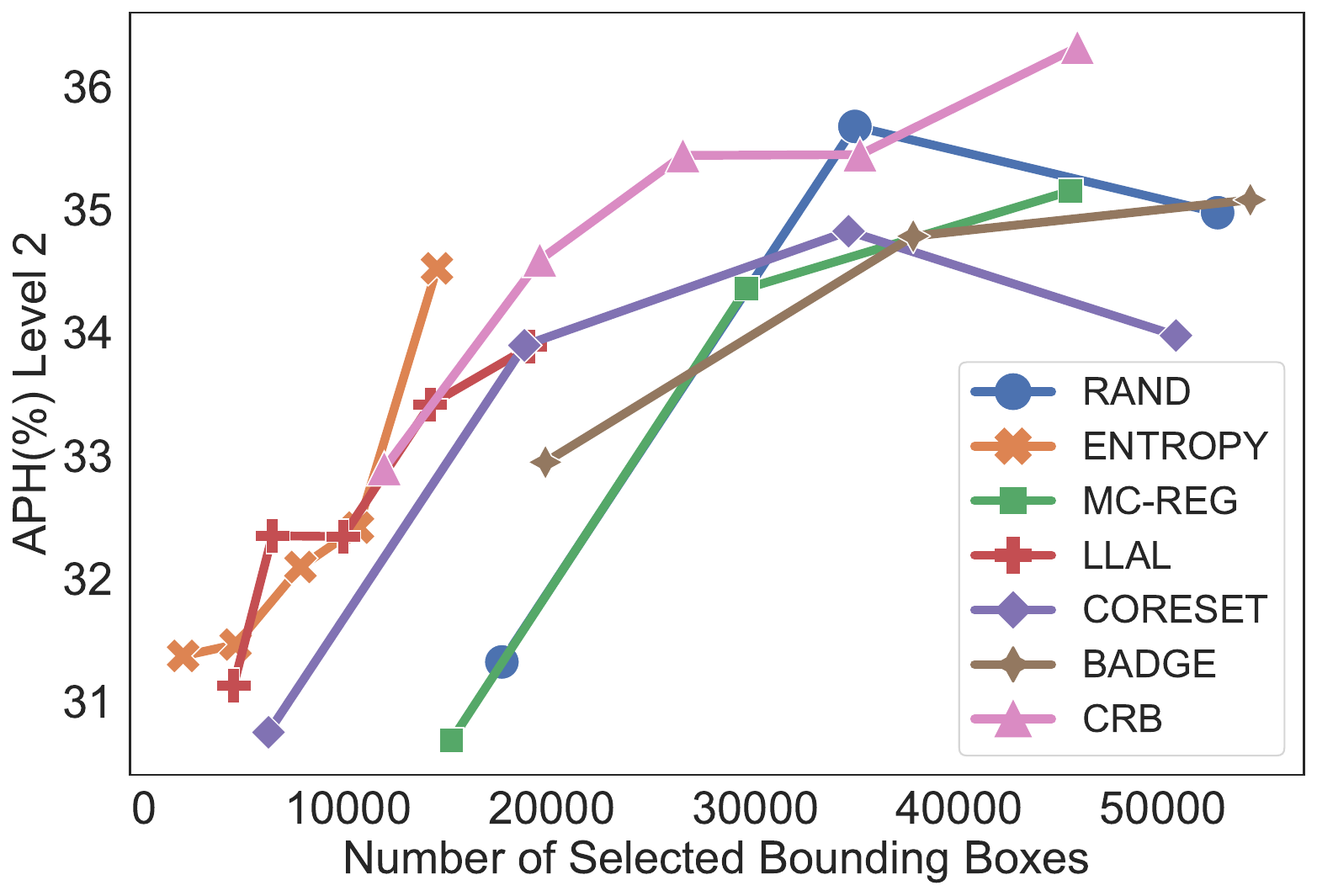} }}%
    \caption{CWAL-3D performance (3D and BEV mAP scores) comparisons of {CRB} and AL baselines on the KITTI and Waymo \textit{val} split, with increasing annototion cost. }%
    \label{fig:kitti_results_boxes}%
\end{figure}

We conducted comprehensive experiments on the KITTI and Waymo datasets with PVRCNN to demonstrate the effectiveness of the proposed CRB approach for the CWAL-3D task. Under a fixed budget of point clouds, the performance of 3D and BEV detection achieved by different AL policies are reported in Figure \ref{fig:kitti_results_boxes}, with standard deviation of three trials shown in shaded regions. We can clearly observe that {CRB} consistently outperforms all state-of-the-art AL methods by a noticeable margin, irrespective of the number of annotated bounding boxes and difficulty settings. It is worth noting that, on the KITTI dataset, the annotation time for the proposed {CRB} is 3 times faster than {Rand}, while achieving a comparable performance. Moreover, AL baselines for regression and classification tasks (\textit{e.g.}, {LLAL}) or for regression only tasks (\textit{e.g.}, {Mc-reg}) generally obtain higher scores yet leading to higher labeling costs than the classification-oriented methods (\textit{e.g.}, {Entropy}). 

Table \ref{tab:generic_applied} reports the major experimental results of the state-of-the-art generic AL methods and applied AL approaches for 2D and 3D detection on the KITTI dataset. It is observed that {LLAL} and {Lt/c} achieve competitive results, as the acquisition criteria adopted jointly consider the classification and regression task. Our proposed {CRB} improves the 3D mAP scores by 6.7\% which validates the effectiveness of minimizing the generalization risk. More qualitative analysis, ablation study and impact of different detector architectures are included in \citep{DBLP:conf/iclr/LuoCWYHB23}.



\begin{table}[t] 
\centering 
\caption{CWAL-3D performance (3D mAP scores) comparisons with generic AL and applied AL for detection on KITTI \textit{val} set with 1\% queried bounding boxes. Moderate difficulty is reported. }
\resizebox{0.9\linewidth}{!}{%
\begin{tabular}{l l c c c c c c}
\toprule 
& Methods &\multicolumn{1}{c}{{Car}}&\multicolumn{1}{c}{{Pedestrian}}&\multicolumn{1}{c}{{Cyclist}} & \multicolumn{1}{c}{{Average}} \\ 
\midrule
\parbox[t]{2mm}{\multirow{3}{*}{\rotatebox[origin=c]{90}{Generic}}} 

&{Coreset} & 77.73 & 41.97 & 59.72 & 59.81 \\
&{Badge} & 75.78 & 46.24 & 62.29 & 61.44 \\
&{LLAL} & 78.65 & 49.87 & 60.35 & 62.95 \\
\midrule
\parbox[t]{2mm}{\multirow{4}{*}{\rotatebox[origin=c]{90}{AL-Det}}} 

& {Mc-reg} & 76.21 & 31.81 & 55.23 & 54.41 \\
& {Mc-mi} & 75.58 & 37.50 & 60.22 & 57.77 \\
& {Consensus} & 78.01 & 49.50 & 55.77 & 61.09 \\
& {Lt/c} & 78.12 & 48.37 & 63.21 & 63.23 \\
\midrule
&{CRB} & \textbf{79.02} & \textbf{54.80} & \textbf{67.45} & \textbf{67.81} \\
\bottomrule 
\end{tabular}
}
\label{tab:generic_applied} 
\end{table}


\section{Discussion and Conclusion}

In this paper, we introduce a novel and realistic problem setting: Open World Active Learning for 3D Object Detection (OWAL-3D), which aims to generalize 3D detectors to open world environments with potential novel classes while minimizing annotation costs. We have developed an extensive open-source benchmark for OWAL-3D, comprising 15 baseline methods and 3 datasets. Additionally, we propose a simple yet highly efficient plug-and-play open world sampling policy, Open Label Conciseness (OLC), and an AL named Open-CRB, specifically tailored for OWAL-3D. Although extensive experiments using our codebase demonstrate the strong performance of Open-CRB, two limitations remain: (1) OLC estimates likelihood of unknown label existence rather than precisely localizing unknown instances, which prevents identifying unknown objects during the test stage before the initial active round selection. (2) OLC primarily addresses semantic shifts (\textit{i.e.}, category mismatches) between pre-trained and test data, but it overlooks covariate shifts within the same classes and scene backgrounds (\textit{e.g.}, adverse weather condition, cross-scene deployment). These limitations highlight the need for future research to leverage the technique from open world object detection to localize unknown instances and domain adaptation approaches to handle both semantic and covariate shifts in open world scenarios.




\balance
\bibliographystyle{IEEEtranN}
\bibliography{bibliography.bib}

\newpage
\begin{IEEEbiography}[{\includegraphics[width=1in,height=1.25in,clip,keepaspectratio,keepaspectratio]{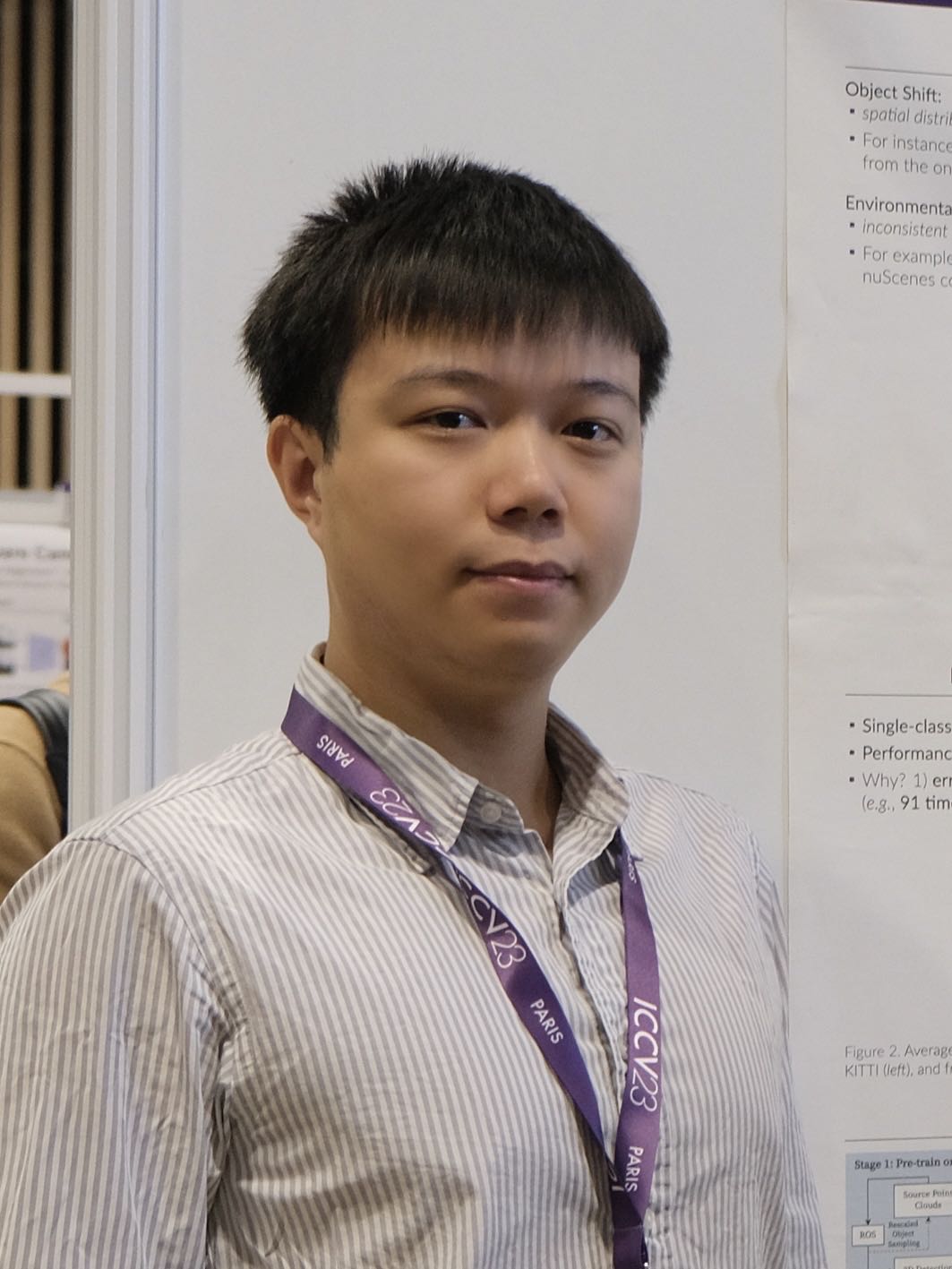}}]{Zhuoxiao Chen} received the bachelor of computer science degree with First Class Honours from the University of Queensland in 2021. He is currently working toward the PhD degree with the University of Queensland. His research interests include 3D computer vision and machine learning.
\end{IEEEbiography}

\begin{IEEEbiography}[{\includegraphics[width=1in,height=1.25in,clip,keepaspectratio,keepaspectratio]{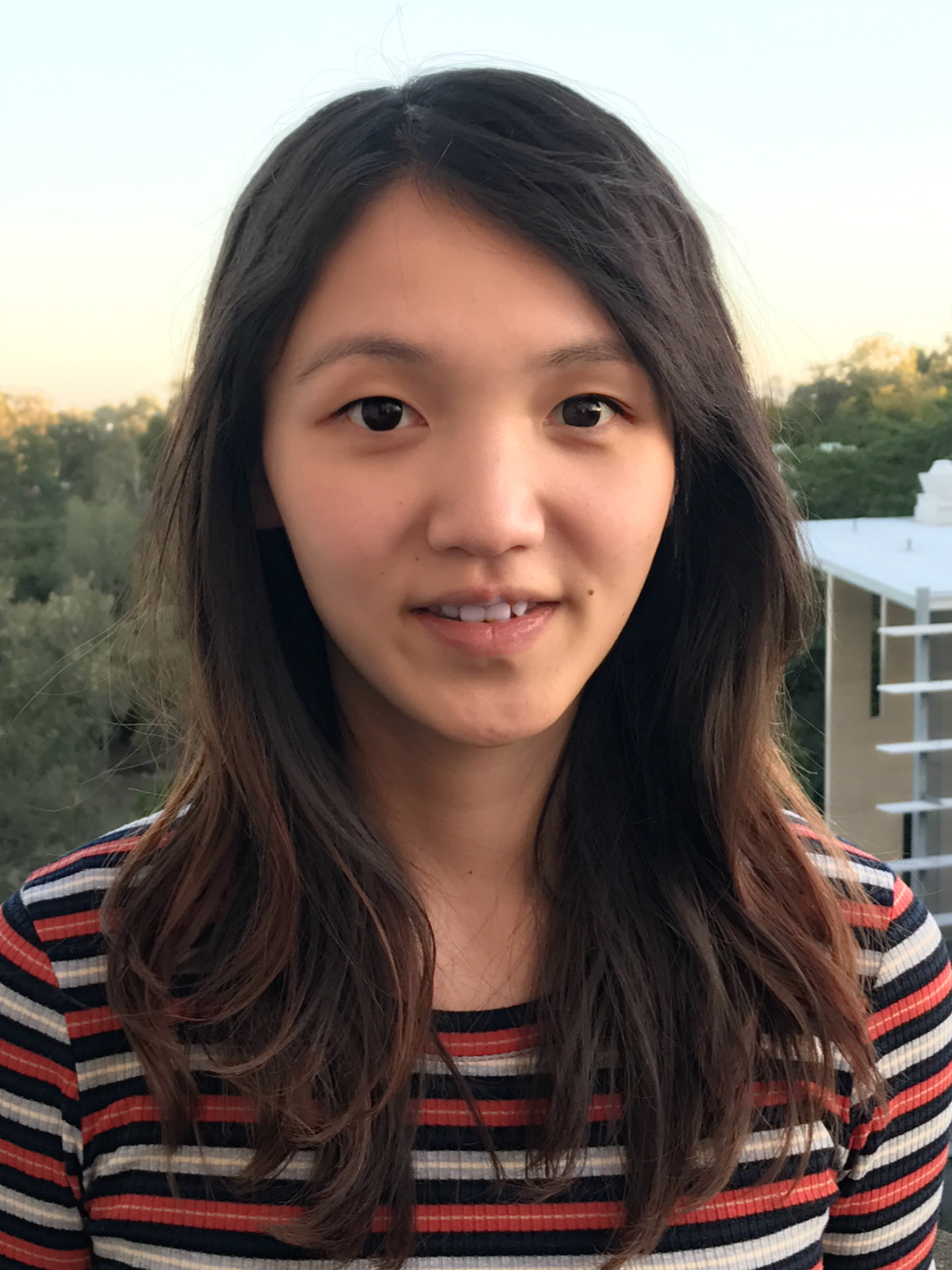}}]{Yadan Luo} (Member, IEEE) received the BS degree in computer science from the University of Electronic Engineering and Technology of China, and the PhD degree from the University of Queensland. Her research interests include machine learning, computer vision, and multimedia data analysis. She is now a lecturer and an ARC DECRA Fellow in the University of Queensland.
\end{IEEEbiography}

\begin{IEEEbiography}[{\includegraphics[width=1in,height=1.25in,clip,keepaspectratio,keepaspectratio]{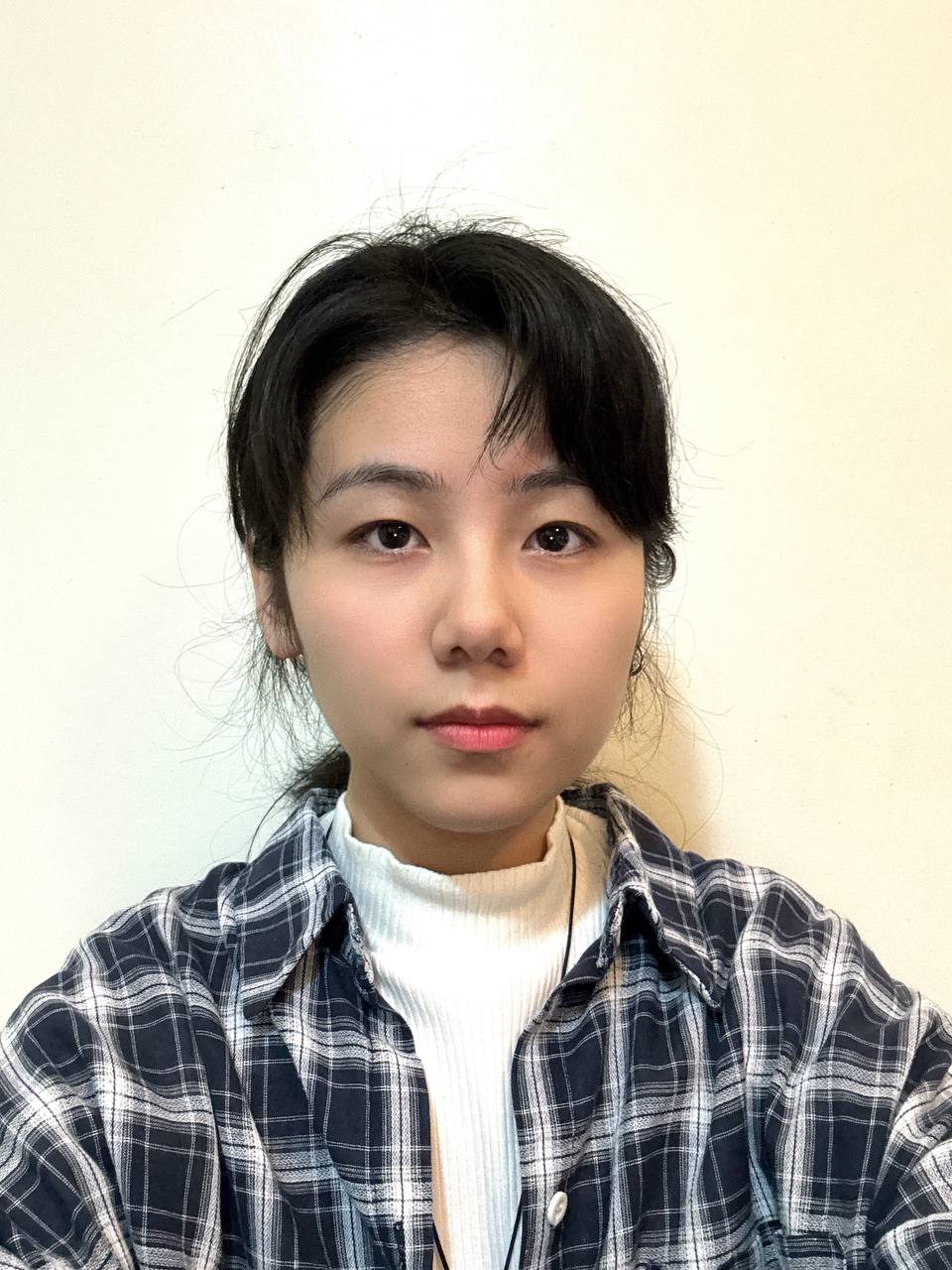}}]{Zixin Wang} received a Bachelor’s degree in management from Shandong Normal University in 2019 and a Master’s degree in IT from the University of Queensland in 2021. She is currently pursuing a PhD at the University of Queensland, with research focusing on computer vision and machine learning.
\end{IEEEbiography}

\begin{IEEEbiography}[{\includegraphics[width=1in,height=1.25in,clip,keepaspectratio]{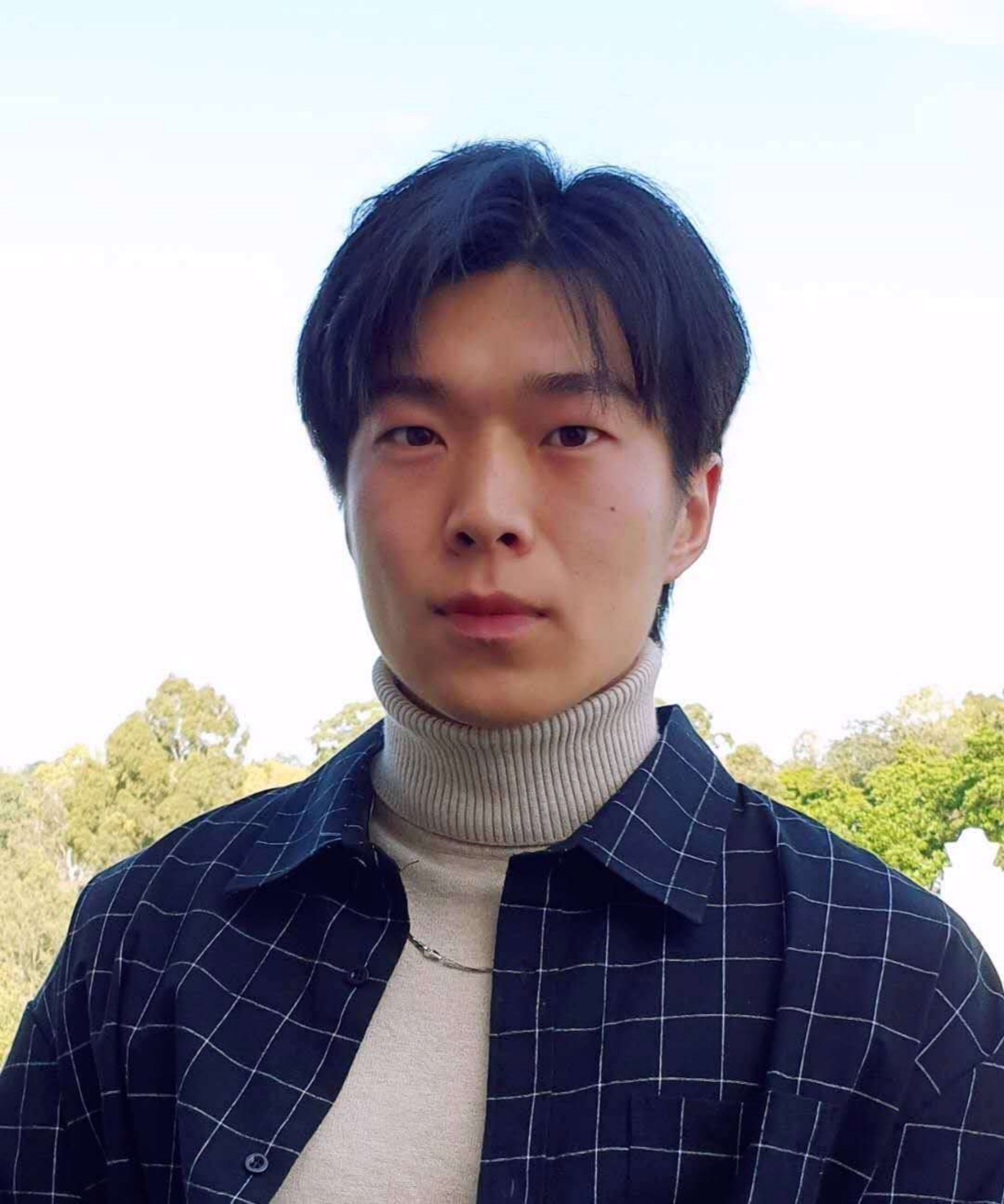}}]{Zijian Wang} received his Ph.D. from the University of Queensland in 2023. His research focuses on model generalization in computer vision. He has published work in top-tier conferences and journals, including ICCV, ICML, ICLR, ACM MM, and TPAMI.
\end{IEEEbiography}
\begin{IEEEbiography}[{\includegraphics[width=1in,height=1.25in,clip,keepaspectratio]{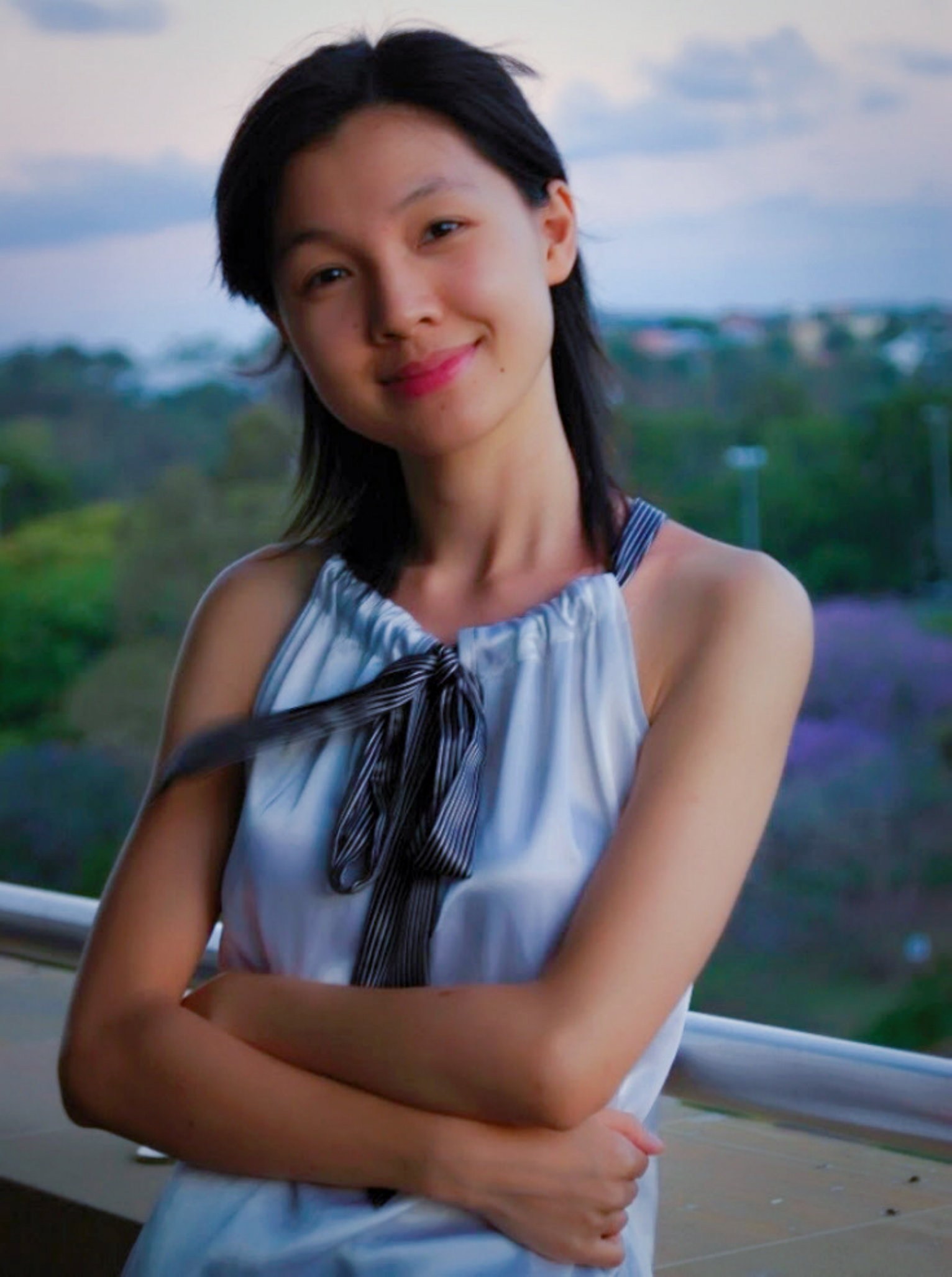}}] {Zi Huang} (Member, IEEE) received the BSc degree from the Department of Computer Science, Tsinghua University, and the PhD degree in computer science from the School of EECS, The University of Queensland, in 2001 and 2007 respectively. She is a professor and Australia Australian Research Council (ARC) Future fellow in the School of EECS, The University of Queensland. Her research interests mainly include Big Data management and analytics, multimedia retrieval and computer vision, and responsible data science. She has served as an associate editor of The VLDB Journal, ACM Transactions on Information Systems, IEEE Transactions on Circuits and Systems for Video Technology, and Pattern Recognition and is a member of the VLDB Endowment Board of Trustees.
\end{IEEEbiography}

\vfill

\end{document}